\documentclass[acmsmall, nonacm]{acmart}
\AtBeginDocument{%
  }

\setcopyright{none}
\renewcommand\footnotetextcopyrightpermission[1]{}

\copyrightyear{2026}
\acmYear{2026 (Preprint, Submitted to ACM TIST)}
\acmDOI{}      
\acmISBN{}     
\acmConference[Under Review]{Work in Progress}{2026}{Taipei, Taiwan}

\acmJournal{TIST}
\acmMonth{4}

\usepackage{amsmath} 
\usepackage[T1]{fontenc}
 
\usepackage{listings}
\usepackage{tcolorbox}
\tcbuselibrary{listings, breakable} 

\newtcblisting{codelist}[1][]{
  breakable,
  colframe=black,
  boxrule=0.6pt,        
  sharp corners,
  enhanced,
  top=2pt, bottom=2pt, left=4pt, right=4pt, 
  listing only,
  listing options={
    upquote=true,
    basicstyle=\ttfamily\scriptsize, 
    breaklines=true,
    columns=fullflexible,
    keepspaces=true,
    lineskip=-0.5pt,
    #1
  }
}

\usepackage[table]{xcolor}
\usepackage{booktabs}   
\usepackage{multirow}   
\usepackage{tabularx}   
\usepackage{amsmath}    

\definecolor{mygreen}{RGB}{60, 180, 100}
\definecolor{freshblue}{RGB}{70, 130, 230}
\definecolor{freshred}{RGB}{235, 90, 90}

\tcbuselibrary{skins}

\begin{document}

\title{CodaRAG: Connecting the Dots with Associativity Inspired by Complementary Learning}




\author{Cheng-Yen Li}
\authornote{Both authors contributed equally to this research.}
\email{r13946003@ntu.edu.tw}
\orcid{0009-0001-3882-9215}
\author{Xuanjun Chen}
\orcid{0000-0002-6042-3633}
\authornotemark[1]
\email{d12942018@ntu.edu.tw}
\author{Claire Lin}
\orcid{0009-0002-6258-1333}
\email{b10705004@ntu.edu.tw}
\author{Wei-Yu Chen}
\orcid{0009-0000-6239-5739}
\email{d14922032@ntu.edu.tw}
\author{Wenhua Nie}
\orcid{0009-0002-7177-038X}
\email{d13944014@ntu.edu.tw}
\author{Hung-yi Lee}
\authornote{Hung-yi Lee is with the NTU Artificial Intelligence Center of Research Excellence, National Taiwan University.}
\orcid{0000-0002-9654-5747}
\email{hungyilee@ntu.edu.tw}
\author{Jyh-Shing Roger Jang}
\orcid{0000-0002-7319-9095}
\email{jang@csie.ntu.edu.tw}
\affiliation{
  \institution{National Taiwan University}
  \country{Taipei, Taiwan}
}

\authorsaddresses{}

\renewcommand{\shortauthors}{C.Y. Li, X. Chen, C. Lin, W.Y. Chen, W. Nie, H. Lee, and J.S.R. Jang}

\begin{abstract}
Large Language Models (LLMs) struggle with knowledge-intensive tasks due to hallucinations and fragmented reasoning over dispersed information. While Retrieval-Augmented Generation (RAG) grounds generation in external sources, existing methods often treat evidence as isolated units, failing to reconstruct the logical chains that connect these dots. Inspired by Complementary Learning Systems (CLS), we propose CodaRAG, a framework that evolves retrieval from passive lookup into active associative discovery. CodaRAG operates via a three-stage pipeline: (1) Knowledge Consolidation to unify fragmented extractions into a stable memory substrate; (2) Associative Navigation to traverse the graph via multi-dimensional pathways—semantic, contextualized, and functional—explicitly recovering dispersed evidence chains; and (3) Interference Elimination to prune hyper-associative noise, ensuring a coherent, high-precision reasoning context. On GraphRAG-Bench, CodaRAG achieves absolute gains of 7–10\% in retrieval recall and 3–11\% in generation accuracy. These results demonstrate CodaRAG’s superior ability to systematically robustify associative evidence retrieval for factual, reasoning, and creative tasks.
\end{abstract}

\ccsdesc[300]{Computing methodologies~Natural language processing}
\ccsdesc[500]{Information systems~Information retrieval}
\ccsdesc[100]{Computing methodologies~Knowledge representation and reasoning}
\keywords{Retrieval-augmented generation, Large language model, Knowledge graph}
\maketitle

\section{Introduction}
Large language models (LLMs) exhibit strong reasoning abilities~\cite{kojima2022zeroshot}, yet their reliance on parametric knowledge limits performance in knowledge-intensive settings, often resulting in hallucinations~\cite{ji2023survey}.
Retrieval-Augmented Generation (RAG)~\cite{lewis2020retrieval} mitigates this issue by grounding generation in externally retrieved evidence.
Prior work shows that RAG’s effectiveness primarily derives from retrieval rather than expanded parametric memory~\cite{izacard2023atlas,ram2023context}, leading to improved factuality and reliability~\cite{es2024ragas,salemi2024evaluating}. However, as the complexity of user queries evolves, the challenge shifts from merely retrieving relevant documents to synthesizing fragmented information across multiple sources \cite{lin2025preliminary}. 
This exposes a fundamental limitation in naive RAG~\cite{gao2023retrieval}, which constructs context as a flat collection of chunks, providing no explicit signal for evidence connectivity and thus limiting reasoning~\cite{mei2025survey}. Graph-based RAG methods introduce structural signals~\cite{peng2025graph,xiang2025use} but remain limited by how the graph is utilized. LightRAG~\cite{guo2024lightrag}, for instance, relies on local one-hop expansion, making it sensitive to entry points and prone to localized noise. Conversely, HippoRAG~\cite{jimenez2024hipporag,gutierrez2025rag} uses the graph primarily for relevance scoring rather than structural organization, failing to assemble retrieved chunks into a coherent evidence chain. Consequently, existing methods may retrieve relevant \emph{dots}, which are atomic units of evidence such as entities and relations, but struggle to bridge them into a coherent context. For example, a query about the iPhone may surface relevant camera specs mixed with unrelated product details, yielding a fragmented context that hampers coherent reasoning. Taken together, these limitations point to the absence of a controlled associative retrieval mechanism that can simultaneously expand to uncover distant associations, organize evidence into a coherent structure, and suppress interfering information. 

To address this gap in organizing retrieved evidence into a coherent context, we draw inspiration from human cognitive architecture. Human cognition excels not merely at storing information, but at actively retrieving and organizing related evidence through associative processes. This perspective is offered by the Complementary Learning Systems (CLS) theory, which characterizes memory as a synergy between a fast, associative hippocampus and a slow, integrative neocortex~\cite{mcclelland1995there,o2014complementary,kumaran2016learning}. By coordinating these dual systems, humans can flexibly navigate complex mental schemas and recover non-obvious connections across fragmented experiences and knowledge ~\cite{kumaran2012generalization,schapiro2017complementary,sun2023organizing}, transforming isolated facts into coherent reasoning chains.
Motivated by this CLS-inspired perspective, we view RAG as a process of \emph{connecting the dots}, by grounding expansive associative discovery within a stable memory substrate.

In this work, we introduce \textbf{CodaRAG}, a RAG framework designed to \emph{Connect the Dots with Associativity}. 
Inspired by CLS~\cite{mcclelland1995there}, CodaRAG augments the retrieval process with associative navigation capabilities, enabling the discovery of logically linked evidence that is often missed by semantic similarity alone.
The framework operates in three stages: 
(1) \emph{Knowledge Consolidation} to \emph{identify the dots}, transforming raw entity extractions into a canonical memory substrate;
(2) \emph{Associative Navigation} to \emph{connect the dots}, coordinating fast semantic linking with contextualized and functional integration; and 
(3) \emph{Interference Elimination} to \emph{refine evidence}, regulating retrieval by suppressing hyper-associative noise. 
These components enable CodaRAG to collect dispersed knowledge into coherent evidence sets for grounded generation.
In our experiments, on GraphRAG-Bench~\cite{xiang2025use}, CodaRAG outperforms HippoRAG 2~\cite{gutierrez2025rag} across multiple domains, yielding 7--10\% higher recall and 3--11\% better accuracy. It demonstrates robust performance on factual, reasoning and creative tasks while balancing the coverage–accuracy trade-off in summarization tasks.

\section{Related Work}
\textbf{Entity Consolidation and Knowledge Grounding.}
While traditional RAG focuses on point-wise vector similarity~\cite{gao2023retrieval}, recent structural approaches, including graph-based and hierarchical methods, introduce explicit organizations to improve global evidence aggregation, such as entity-relation graphs in GraphRAG~\cite{edge2024local} and LightRAG~\cite{guo2024lightrag}, as well as hierarchical text representations in RAPTOR~\cite{sarthi2024raptor}.
Despite these advances, existing graph-based methods often operate directly on surface-level mentions without explicit type abstraction, which can result in inconsistent extraction of semantically similar entities. In addition, they fail to resolve heterogeneous forms into unified entity representations, resulting in fragmented graph structures~\cite{shen2014entity,hogan2021knowledge}.
CodaRAG resolves this through \emph{Knowledge Consolidation}, which guides structured extraction and merges redundant entities into a unified memory substrate for associative retrieval.

\textbf{Multi-dimensional Associative Navigation.} 
Effective multi-hop reasoning requires navigating relational paths that extend beyond surface-level semantic similarity~\cite{tang2024multihop,peng2025graph}.
Cognitive-inspired frameworks such as HippoRAG~\cite{jimenez2024hipporag,gutierrez2025rag} employ Personalized PageRank (PPR)~\cite{haveliwala2002topic} to propagate query influence over the graph and select relevant text chunks.
However, reliance on a singular associative mechanism may constrain the ability to capture diverse relational patterns across heterogeneous evidence.
Drawing on CLS theory~\cite{mcclelland1995there}, CodaRAG moves beyond single-mode search by integrating \emph{semantic, contextualized, and functional} pathways, ensuring that non-obvious evidence is captured through this multidimensional navigation.

\textbf{Interference Regulation and Conflict Resolution.}
Multi-hop associative navigation inevitably introduces informational interference~\cite{liu2025hoprag}, 
such noise accumulation can lead to evidence degradation, ultimately compromising the integrity of the reasoning process~\cite{yoran2023making}.
While expansive graph traversal can improve recall~\cite{peng2025graph}, it also risks hyper-associativity, where loosely constrained associations obscure the reasoning path. 
Despite the risk, most graph-based RAG methods lack explicit mechanisms to regulate such interference during retrieval. To address this, CodaRAG implements \emph{Interference Elimination}, acting as a cognitive filter that prunes hyper-associative noise and resolves evidence conflicts to ensure a clean, high-precision context for reasoning. 

\begin{figure*}[!t]
  \centering
    \includegraphics[width=1\textwidth]{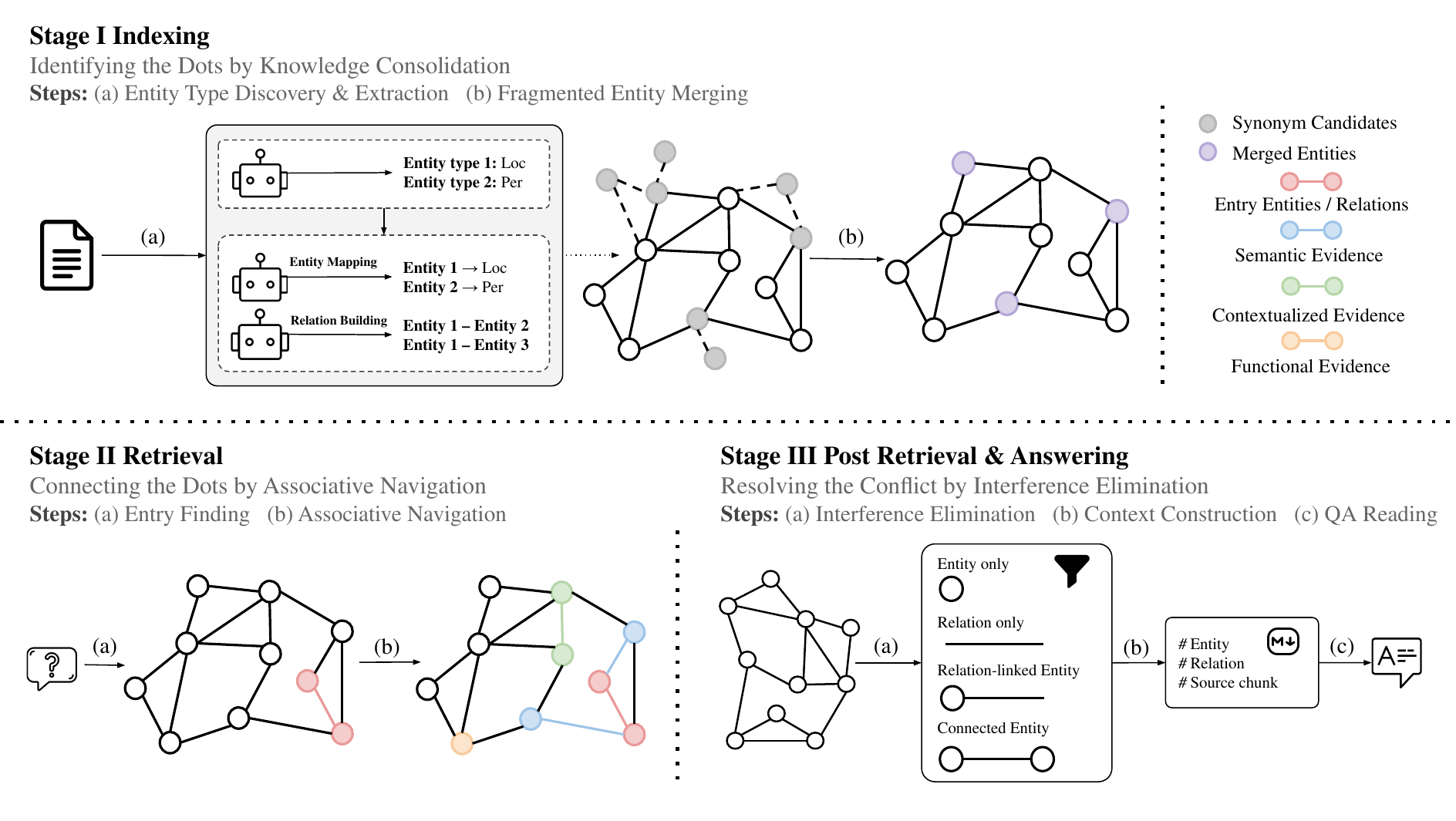}
    \vspace{-2.em}
    \caption{
    The CodaRAG Framework: Connecting the Dots with Associativity. The workflow begins by transforming unstructured text into a consolidated KG (Stage I), which is then traversed via multi-dimensional Associative Navigation (Stage II) to uncover semantic, contextualized, and functional evidence. Finally, Stage III applies Interference Elimination to filter out noise, ensuring only the most relevant context is used for the LLM-generated answer.
    }
    \Description{
Overview of the CodaRAG framework as a three-stage pipeline. 
Stage I performs knowledge consolidation, including Entity Type Discovery and Fragmented Entity Merging, transforming unstructured text into a structured knowledge graph. 
Stage II retrieves evidence by first identifying entry nodes and then applying Associative Navigation to expand into semantic, contextualized, and functional evidence subgraphs. 
Stage III filters irrelevant entities and relations, constructs the final context, and generates answers using a language model.
}
  \label{fig:overview_framework}
\end{figure*}

\section{The CodaRAG Framework}

To bridge the \emph{connectivity gap}, the inability of vector systems to link scattered evidence, CodaRAG transitions from a flat retrieval to an associative navigation based framework via three stages:
\textbf{1) Identifying the Dots:} 
We perform Knowledge Consolidation to unify fragmented mentions into a canonical memory substrate, providing a stable foundation for global navigation. 
\textbf{2) Connecting the Dots:} Recognizing that human-like reasoning relies on more than just semantic similarity, we utilize complementary pathways (semantic, contextual, and functional) to recover hidden links invisible to rigid, single-metric strategies.
\textbf{3) Regulating Interference:} We mitigate the risk of \emph{hyper-associativity} through an executive control mechanism, filtering out noise into coherent, task-relevant evidence.
In the following sections, we first formalize the problem (Section~\ref{sub:overview}), and then detail the pipeline: Knowledge Consolidation (Section~\ref{subsec:knc}), Associative Navigation (Section~\ref{sec:navigation}), and Interference Elimination (Section~\ref{sec:inference}).

\subsection{Problem Formulation and Overview}\label{sub:overview}
Given a query \( q \) and a document corpus, CodaRAG synthesizes scattered evidence into a compact and coherent context for grounded answer generation.
The system transforms the corpus into a consolidated KG \( G = (E, R) \), where \( e \in E \) denotes an entity node and \( r \in R \) a relation edge, which together constitute the system’s memory. 
The entities and relations within the KG are referred to as \emph{dots}, which serve as the atomic memory units for retrieval and grounding.
Retrieval aims to extract a query-relevant evidence subgraph \( G_q \subseteq G \) that provides sufficient support for the answer while filtering out irrelevant or redundant information that may interfere with generation.
This objective aims to increase evidentiary coverage of the subgraph while reducing informational interference that may destabilize the generative process.
%
As shown in Figure~\ref{fig:overview_framework}, CodaRAG follows a three-stage design.
At indexing time, \emph{Knowledge Consolidation} transforms raw extractions into a unified KG.
At retrieval time, \emph{Associative Navigation} takes the query and the consolidated KG as input, and produces an evidence subgraph via complementary associations.
Finally, \emph{Interference Elimination} filters this subgraph into a compact evidence set, which is then grounded in supporting text chunks for generation.

\begin{figure*}[t]
    \centering
    \includegraphics[width=1\textwidth]{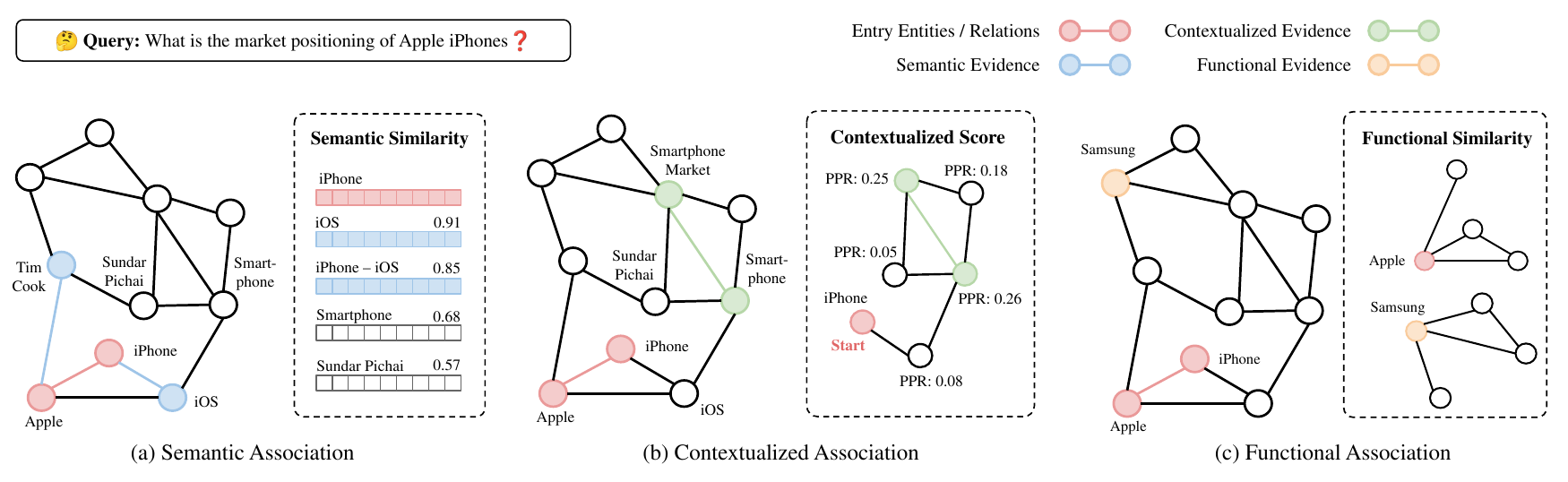}
    \vspace{-2em}
    \caption{
      Overview of complementary Associative Navigation strategies. The CodaRAG framework enhances retrieval by integrating three core mechanisms: (a) Semantic Association for local entity expansion via semantic similarity; (b) Contextualized Association for query-aware importance propagation; and (c) Functional Association for identifying structurally analogous entities through topology analysis. Together, these strategies enable multi-dimensional navigation to connect disparate but relevant information. 
    }
    \Description{Illustration of three Associative Navigation pathways over a knowledge graph given a query. Starting from entry entities, (a) Semantic Association expands to nearby nodes based on similarity scores, 
    (b) Contextualized Association propagates query-aware importance using scores, and 
    (c) Functional Association retrieves structurally analogous nodes based on graph topology. 
    Each pathway produces a distinct evidence subgraph for retrieval.
    }
    \label{fig:navigation}
\end{figure*}

\subsection{Identifying the Dots by Knowledge Consolidation}
\label{subsec:knc}
To connect the dots, they must first be identified within a unified semantic space. CodaRAG transforms indexing to consolidate raw extractions into a structured memory substrate, bridging categorical gaps and entity fragmentation. 
Through Entity Type Discovery and Extraction, coupled with Fragmented Entity Merging, we construct a canonical foundation that yields a consistent and navigable structure rather than disjointed mentions.

\textbf{Entity Type Discovery and Extraction.}
A key barrier to connecting dots lies in the absence of appropriate entity typing guidance across domains, which can result in inconsistent extraction of semantically similar entities, with some instances being identified while others are omitted. This issue is further exacerbated by current extraction pipelines: Closed Information Extraction (ClosedIE)~\cite{ratinov2009design} imposes overly restrictive schemas, while Open Information Extraction (OpenIE)~\cite{etzioni2008open} avoids explicit typing, leaving the underlying categorical structure largely unmapped.

To address this, we perform Entity Type Discovery on top of extraction, inducing semantic categories directly from corpus-level patterns without assuming a fixed schema.
Inspired by theories of semantic cognition~\cite{mcclelland2003parallel}, we adopt a \emph{suggest--refine} paradigm: document-level contexts first propose candidate types during extraction, which are then aggregated and consolidated across documents into a shared type inventory.
For example, inducing an \textit{organization} type enables mentions such as \emph{Google}, \emph{Microsoft}, and \emph{OpenAI} to be consistently aligned across diverse contexts.
This induced semantic scaffold facilitates type-aware alignment and improves consistency in graph construction.
The decision logic and prompt configurations are detailed in Appendix~\ref{sec:discovery_prompt}.

\textbf{Fragmented Entity Merging.}
Another barrier to connectivity arises in multi-chunk or multi-document settings, where a single real-world concept is mentioned across different passages using varied surface forms and partial descriptions.
When treated as isolated entities, such fragmentation introduces structural discontinuities that obstruct latent logical pathways and stifle associative navigation across the graph. 

Fragmented Entity Merging mitigates this issue by selectively integrating highly similar entity representations during indexing.
Inspired by theories of memory integration~\cite{schlichting2015memory}, which highlight selective memory consolidation, we first identify candidate entity pairs based on embedding-level similarity, and employ a similarity-based gate that forwards only pairs exceeding a predefined threshold to an LLM-based judge for final merging decisions, during which duplicated relations are merged into a single edge.
For example, mentions such as \emph{Google}, \emph{Google Inc.}, and \emph{Alphabet's Google} appearing in different documents or chunks can be consolidated into a single \textit{Google} entity node.
This fusion reduces redundancy, strengthens relational consistency, and repairs structural gaps in the graph, restoring traversable latent connectivity essential for associative retrieval.
The decision logic and prompts are defined in the Appendix~\ref{sec:merging_prompt}.

Our indexing pipeline builds on LightRAG~\cite{guo2024lightrag}, which organizes documents into entity-relation graphs with descriptive metadata.
CodaRAG frames indexing as a consolidation process.
Through Entity Type Discovery and Fragmented Entity Merging, we uncover latent connections that better support associative navigation.

\subsection{Connecting the Dots by Associative Navigation}
\label{sec:navigation} 
Following the identification and refinement of the dots, CodaRAG bridges them via Associative Navigation, an active mechanism mimicking the brain’s spreading activation.
Guided by a CLS-inspired perspective, we abstract its core intuition of complementary processing and instantiate it as multiple retrieval pathways—Semantic, Contextualized, and Functional association—each capturing different aspects of evidence.
As illustrated in Fig.~\ref{fig:navigation}, given the query \textit{``What is the market positioning of Apple iPhones?''}, the system first performs Entry Finding to identify query-relevant entities and relations (e.g., \textit{Apple}, \textit{iPhone}).
Semantic Association retrieves locally connected entities (e.g., \textit{iOS}, \textit{Tim Cook}),
Contextualized Association captures globally salient concepts (e.g., \textit{smartphone market}),
while Functional Association identifies structurally analogous entities (e.g., \textit{Samsung}).
By integrating complementary evidence from distinct retrieval pathways, CodaRAG links dispersed evidence into a more coherent evidence subgraph.

\textbf{Entry Finding.}
To initialize Associative Navigation, we identify a set of anchor dots in the KG, which serve as entry points for subsequent spreading activation.
Following LightRAG~\cite{guo2024lightrag}, an LLM transforms the query $q$ into low- and high-level query-related cues, which are matched against entities and relations to select entry dots.
These anchors serve as the starting points from which we coordinate complementary associative processes. 
To quantify these associations, we define three complementary scoring functions capturing local, global, and structural proximity.

\textbf{Semantic Association Linking.}
Semantic Association reflects a fast, local retrieval behavior inspired by CLS~\cite{mcclelland1995there} by performing entry-centric exploration over the KG.
Starting from the entry entities, it expands locally to retrieve entities and relations that are both topologically close and semantically aligned with the query.
Through this neighborhood-centric expansion, Semantic Association enables access to concrete and explicitly related evidence.
For a candidate entity $e$ connected to an entry entity $e_0$ through a connecting relation $r$, we define the Semantic Association score as:
\begin{equation}
\label{eq:sea_score}
S_{\text{SemA}}(e)
=
\alpha\, \mathrm{sim}_{\text{rel}}(r,q)
+
\beta\, \mathrm{sim}_{\text{ent}}(e,q).
\end{equation}
where $\alpha$ and $\beta$ are scalar coefficients that control the contribution of relation-level and entity-level similarity.
Here, $\mathrm{sim}_{\text{rel}}$ measures the similarity between high-level query-related cues and the connecting relation, while $\mathrm{sim}_{\text{ent}}$ measures the similarity between low-level cues and the candidate entity. 
We prune connections below the threshold $\tau$ and retain only the top-$k$ neighbors per entity.


\textbf{Contextualized Association Bridging.}
Contextualized Association captures a slower, more integrative retrieval behavior, emphasizing global contextual integration.
It propagates query-conditioned relevance from the entry entities over the KG using PPR~\cite{haveliwala2002topic}, resulting a set of globally salient top-$k$ entities; relations are then induced from the internal connections among these selected entities to form a contextualized evidence subgraph.
Formally, Contextualized Association estimates global entity importance through a propagation update:

\begin{equation}
\label{eq:coa_ppr_coa_message}
S_{\text{ConA}}(e)
=
(1-d)\,p_{E_0}(e)
+
d\, M_q(e),
\quad
\text{where}
\quad
M_q(e)
=
\sum_{e' \in \mathcal{N}(e)}
S_{\text{ConA}}(e')\, p_q(e \mid e').
\end{equation}
Equations~(\ref{eq:coa_ppr_coa_message}) define an iterative propagation process, where entity scores are updated via message passing over the graph until convergence or for a fixed number of steps.
Here, $\mathcal{N}(e)$ denotes the set of neighboring entities of $e$. 
The message term $M_q(e)$ is defined by a softmax-normalized edge transition distribution $p_q(e \mid e')$, initialized from relations aligned with high-level query-related cues. 
The personalization term $p_{E_0}(e)$ is a query-conditioned probability distribution defined over the set of entry entities $E_0$, which is initialized using a softmax-normalized similarity score derived from low-level query-related cues, thereby reflecting their relative relevance to the given query.

\textbf{Functional Association Mapping.}
Functional Association further extends this integrative process by capturing higher-level structural roles and relational patterns in KG.
It operates on structural embeddings of entities computed from the KG using Fast Random Projection (FastRP)~\cite{chen2019fast}, producing a set of top-$k$ functionally analogous entities together with the relational structure induced among them in the KG. Details are provided in Appendix~\ref{sec:appendix_hyperparameters}. 
%
Formally, let $\mathbf{x}_e \in \mathbb{R}^d$ denote the structural embedding of entity $e$, with all entities embedded in the same vector space.
Given a set of entry entities $E_0$, each associated with an embedding $\mathbf{x}_{e_0}$, the score for an entity $e$ is defined as the average structural similarity to the entry set $E_0$:
\begin{equation}
\label{eq:sta_score}
S_{\text{FunA}}(e)
=
\frac{1}{|E_0|}
\sum_{e_0 \in E_0}
\mathrm{sim}
\bigl(
\mathbf{x}_{e_0},\,
\mathbf{x}_e
\bigr).
\end{equation}
The structural embeddings are computed over a weighted KG, where each relation edge is assigned a weight based on its corpus-level co-occurrence frequency

\subsection{Resolving the Conflict by Interference Elimination} \label{sec:inference}
While Associative Navigation increases coverage, it also introduces the risk of \emph{associative saturation}, where an overabundance of loosely connected dots leads to increased semantic ambiguity. Inspired by prefrontal executive control~\cite{miller2001integrative}, we introduce an \textit{Interference Elimination} stage that serves as a cognitive filtering mechanism. This stage evaluates retrieved entities and relations against the given query and removes associations that do not meaningfully contribute to downstream generation. The detailed decision logic and prompt design are provided in Appendix~\ref{sec:IE_prompt}.
Finally, the resulting entities and relations are aligned with their corresponding supporting top-$k$ source chunks, which are ranked based on the frequency of entity and relation occurrences, to construct the final context.

\section{Experimental Setup}
\label{sec:experimental_setup}
\textbf{Benchmark.}
We evaluate our framework on GraphRAG-Bench~\cite{xiang2025use}\footnote{
GraphRAG-
Bench: \href{https://github.com/GraphRAG-Bench/GraphRAG-Benchmark}{github.com/GraphRAG-Bench/GraphRAG-Benchmark}},
which is explicitly designed for graph-based RAG.
Compared to general-purpose benchmarks, it more directly reflects the effects of graph construction and structured retrieval.
Due to resource constraints, we evaluate on sampled questions from two subsets across four task categories:
Fact Retrieval (Fact), Complex Reasoning (Reason), Contextual Summarization (Summary), and Creative Generation (Creation).
The Medical subset consists of structured clinical data and uses a balanced sample of 50 questions per task,
while the Novel subset comprises unstructured narratives and is sampled at the document level,
resulting in 273, 163, 98, and 18 questions for Fact, Reason, Summary, and Creation, drawn from six documents.


\textbf{Evaluation Metrics.}
We adopt the multi-dimensional evaluation framework of GraphRAG-Bench~\cite{xiang2025use}.
Retrieval is evaluated using context relevance (REL) and evidence recall (REC),
while generation quality is assessed with Rouge-L (RGL)~\cite{lin2004rouge},
accuracy (ACC), faithfulness (FTH), and coverage (COV).
Except for Rouge-L, these metrics involve LLM-based evaluation components as defined in the benchmark.
Metric definitions are provided in Appendix~\ref{sec:eval_metrics}.

\textbf{Implementation Details.} 
We use GPT-5-mini~\cite{openai_gpt5mini_2025} as the backbone model for both indexing and generation, aligning with prior  work~\cite{guo2024lightrag,gutierrez2025rag,xiang2025use}, which adopts GPT-family models for fair comparison.
Entities, relations, and text chunks are encoded using text-embedding-3-large~\cite{openai_text_embedding_2024}, balancing embedding quality and indexing throughput.
For \emph{Semantic Association}, relation-level weights ($\alpha=0.5$) are set higher than entity-level weights ($\beta=0.4$), as relations directly capture source-to-target connectivity, and connections with scores below $\tau=0.35$ are pruned. These coefficients and thresholds are selected via a coarse grid search on a held-out subset of the data, based on retrieval and generation metrics, and then fixed for all subsequent experiments.
For \emph{Contextualized Association}, we adopt PPR~\cite{haveliwala2002topic} with damping factor $d=0.85$.
All hyperparameter settings used in our experiments are summarized in Appendix~\ref{sec:appendix_hyperparameters}.

\begin{table*}[t]
  \caption{Performance comparison across diverse retrieval and generation tasks. 
  }
  \vspace{-1em}
  \label{tab:unified_perf}
  \centering
  {\fontsize{7.5}{9}\selectfont
  \setlength{\tabcolsep}{3.5pt}
  \renewcommand{\arraystretch}{1.15}
  \resizebox{\linewidth}{!}{
    \begin{tabular}{lcccccccc|ccccccccc}
      \toprule
      & \multicolumn{8}{c}{\textbf{Retrieval Performance $\uparrow$}}
      & \multicolumn{9}{c}{\textbf{Generation Performance $\uparrow$}} \\
      \cmidrule(lr){2-9}\cmidrule(lr){10-18}
      Method
      & \multicolumn{2}{c}{Fact}
      & \multicolumn{2}{c}{Reason}
      & \multicolumn{2}{c}{Summary}
      & \multicolumn{2}{c}{Creation}
      & \multicolumn{2}{c}{Fact}
      & \multicolumn{2}{c}{Reason}
      & \multicolumn{2}{c}{Summary}
      & \multicolumn{3}{c}{Creation} \\
      \cmidrule(lr){2-3} \cmidrule(lr){4-5} \cmidrule(lr){6-7} \cmidrule(lr){8-9} \cmidrule(lr){10-11} \cmidrule(lr){12-13} \cmidrule(lr){14-15} \cmidrule(lr){16-18}
      & REL & REC
      & REL & REC
      & REL & REC
      & REL & REC
      & RGL & ACC
      & RGL & ACC
      & COV & ACC
      & COV & FTH & ACC \\
      \midrule
      \multicolumn{18}{l}{\textbf{Medical Dataset}} \\
      LightRAG
      & 91.00 & 90.65
      & 94.00 & 89.61
      & 93.00 & 94.06
      & 81.50 & 65.32
      & 12.52 & 49.04
      & 8.63 & 53.43
      & \textbf{90.17} & 63.61
      & 52.40 & 71.51 & 62.16 \\
      HippoRAG~2
      & 89.50 & 90.83
      & 83.50 & 88.28
      & 91.00 & 93.75
      & 77.00 & 66.34
      & 36.20 & 68.37
      & \textbf{27.17} & 64.76
      & 68.60 & \textbf{77.69}
      & 55.75 & 50.98 & 50.89 \\
      \rowcolor{blue!5} Ours (KG-only)
      & \textbf{96.00} & 92.33
      & \textbf{98.00} & 87.67
      & \textbf{98.50} & 91.73
      & 82.50 & 71.22
      & 36.03 & 68.35
      & 24.06 & 62.36
      & 77.46 & 67.48
      & 54.95 & \textbf{85.97} & 64.72 \\
      \rowcolor{blue!5} Ours (KG + Chunks)
      & \textbf{96.00} & \textbf{95.67}
      & \textbf{98.00} & \textbf{90.43}
      & 97.00 & \textbf{96.89}
      & \textbf{86.00} & \textbf{75.77}
      & \textbf{42.25} & \textbf{71.94}
      & 25.74 & \textbf{67.36}
      & 81.70 & 73.11
      & \textbf{59.50} & 84.56 & \textbf{66.78} \\
      \midrule
      \multicolumn{18}{l}{\textbf{Novel Dataset}} \\
      LightRAG
      & 96.71 & 90.29
      & 94.17 & 89.20
      & 94.46 & 90.81
      & 93.22 & 59.75
      & 18.73 & 51.35
      & 9.79 & 40.89
      & \textbf{75.11} & 47.02
      & 45.45 & 63.35 & 36.52 \\
      HippoRAG~2
      & 86.63 & 78.21
      & 85.58 & 78.81
      & 89.29 & 80.75
      & 88.89 & 55.88
      & 31.91 & 60.38
      & 24.35 & 47.31
      & 57.21 & \textbf{51.32}
      & \textbf{50.86} & 43.80 & 39.77 \\
      \rowcolor{blue!5} Ours (KG-only)
      & 95.97 & 84.68
      & 91.10 & 81.70
      & 94.90 & 82.72
      & 90.10 & 50.73
      & \textbf{41.52} & 60.60
      & \textbf{25.67} & 47.69
      & 58.94 & 48.88
      & 40.18 & \textbf{75.75} & 43.78 \\
      \rowcolor{blue!5} Ours (KG + Chunks)
      & \textbf{99.08} & \textbf{94.07}
      & \textbf{95.56} & \textbf{91.44}
      & \textbf{97.96} & \textbf{91.27}
      & \textbf{95.89} & \textbf{60.20}
      & 40.91 & \textbf{62.87}
      & 25.40 & \textbf{49.87}
      & 66.11 & 50.50
      & 48.60 & 73.65 & \textbf{46.79} \\
      \midrule
      \multicolumn{18}{l}{\scriptsize Note: Evaluation metric definitions are detailed in Sec.~\ref{sec:experimental_setup}. The symbol $\uparrow$ indicates that higher scores represent better performance.} \\
    \end{tabular}
}}

\end{table*}

\section{Main Results}

We benchmark against two representative graph-based RAG baselines: LightRAG~\cite{guo2024lightrag} \footnote{LightRAG: \href{https://github.com/HKUDS/LightRAG}{github.com/HKUDS/LightRAG}}
and HippoRAG~2~\cite{gutierrez2025rag}
\footnote{HippoRAG~2: \href{https://github.com/OSU-NLP-Group/HippoRAG}{github.com/OSU-NLP-Group/HippoRAG}}. 
The former serves as our implementation foundation, while the latter represents the strongest reported baseline on GraphRAG-Bench~\cite{xiang2025use}. 
We additionally include \texttt{KG-only} as a dedicated setting to isolate and analyze the effect of structured graph retrieval independently from textual augmentation.
Table~\ref{tab:unified_perf} reports retrieval and generation performance.

\textbf{Baseline Comparison.}
We conduct a comparative evaluation of performance across different graph-based RAG systems. As shown in Table~\ref{tab:unified_perf}, our full system (\texttt{KG + Chunks}) consistently achieves the strongest relevance--recall trade-off across all question types. Compared to the strong baseline HippoRAG~2~\cite{gutierrez2025rag}, CodaRAG demonstrates consistent average absolute gains of 7--10\% in retrieval recall (REC) and 3--11\% in generation accuracy (ACC) across Fact, Reason, and Creation tasks on both the Medical and Novel datasets.

For Summary tasks, our method exhibits a clear coverage--accuracy trade-off, favoring broader evidence coverage (COV) while retaining competitive accuracy (ACC), whereas baseline methods tend to achieve higher accuracy (ACC) at the expense of limited coverage (COV) or, conversely, improve coverage (COV) while introducing noisier and less precise information. 
On Creation queries, it yields better-grounded responses with improved faithfulness (FTH), highlighting CodaRAG’s ability to synthesize dispersed evidence. Overall, our method achieves strong performance in both retrieval and generation, outperforming LightRAG~\cite{guo2024lightrag} and HippoRAG~2~\cite{gutierrez2025rag} across most settings.

\textbf{The Complementarity of KG and Chunks.}
Comparing \texttt{KG-only} and \texttt{KG + Chunks} reveals a clear complementarity between structured graph grounding and textual augmentation.
Notably, the \texttt{KG-only} setting already achieves strong retrieval and generation performance, indicating that the consolidated KG provides a solid  foundation. While the \texttt{KG-only} setting favors higher faithfulness (FTH), incorporating text chunks improves coverage (COV) and accuracy (ACC), albeit with a slight reduction in faithfulness due to the introduction of additional contextual noise.
Together, these results suggest that graph-based retrieval, when augmented with text chunks, enables a balanced trade-off between grounding reliability and evidence coverage.
A more detailed component-level analysis is provided in Section~\ref{sec:ablation}.

\begin{table*}[t]
  \caption{
    Ablation analysis of core components on the Medical dataset. 
    We conduct ablation studies by comparing our full framework with five variants: (1) Base (the original LightRAG~\cite{guo2024lightrag} baseline); (2) w/o SemA, w/o ConA, and w/o FunA, stripping the Semantic, Contextualized, and Functional Association modules; and (3) w/o Elim, excluding the Interference Elimination module. }
  \vspace{-1em}
  \label{tab:ablation_all}
  \centering
  {\fontsize{7.8}{10}\selectfont
  \setlength{\tabcolsep}{4pt}
  \renewcommand{\arraystretch}{1.15}
  \resizebox{\linewidth}{!}{
    \begin{tabular}{lcccccccc|ccccccccc}
      \toprule
      & \multicolumn{8}{c}{\textbf{Retrieval Performance $\uparrow$}}
      & \multicolumn{9}{c}{\textbf{Generation  Performance $\uparrow$}} \\
      \cmidrule(lr){2-9}\cmidrule(lr){10-18}
      Method
      & \multicolumn{2}{c}{Fact}
      & \multicolumn{2}{c}{Reason}
      & \multicolumn{2}{c}{Summary}
      & \multicolumn{2}{c}{Creation}
      & \multicolumn{2}{c}{Fact}
      & \multicolumn{2}{c}{Reason}
      & \multicolumn{2}{c}{Summary}
      & \multicolumn{3}{c}{Creation} \\
      \cmidrule(lr){2-3} \cmidrule(lr){4-5} \cmidrule(lr){6-7} \cmidrule(lr){8-9} \cmidrule(lr){10-11} \cmidrule(lr){12-13} \cmidrule(lr){14-15} \cmidrule(lr){16-18}
      & REL & REC
      & REL & REC
      & REL & REC
      & REL & REC
      & RGL & ACC
      & RGL & ACC
      & COV & ACC
      & COV & FTH & ACC \\
      \midrule
      \multicolumn{18}{l}{\textbf{Stage I Indexing:} Identifying the Dots by Knowledge Consolidation} \\
      Base
      & \textbf{94.50} & 73.65
      & 89.00 & 66.58
      & 79.00 & 66.40
      & 52.50 & 35.84
      & \textbf{30.67} & 56.08
      & 24.97 & 60.41
      & 52.50 & 63.23
      & 32.64 & 76.97 & 59.30 \\
      \rowcolor{blue!5} Ours
      & 91.00 & \textbf{80.78}
      & \textbf{92.50} & \textbf{70.11}
      & \textbf{83.00} & \textbf{68.28}
      & \textbf{58.50} & \textbf{40.38}
      & 30.31 & \textbf{57.20}
      & \textbf{25.28} & \textbf{60.54}
      & \textbf{60.04} & \textbf{66.12}
      & \textbf{37.63} & \textbf{82.46} & \textbf{61.33} \\
      \midrule
      \multicolumn{18}{l}{\textbf{Stage II Retrieval:} Connecting the Dots by Associative Navigation} \\
      w/o SemA
      & 94.00 & 86.35
      & 92.00 & 79.81
      & 88.00 & 77.64
      & 63.00 & 57.38
      & 33.76 & 60.30
      & 26.17 & 62.06
      & 61.95 & 64.81
      & 46.48 & 84.93 & 65.77 \\
      w/o ConA
      & 95.50 & 85.95
      & 92.00 & 80.40
      & 93.50 & 85.98
      & 71.00 & 60.17
      & 34.64 & 64.51
      & 25.80 & 62.85
      & 72.01 & 67.50
      & 49.39 & 82.72 & \textbf{66.81} \\
      w/o FunA
      & 94.00 & 87.03
      & \textbf{97.00} & 81.80
      & \textbf{94.00} & 85.68
      & \textbf{74.00} & \textbf{64.03}
      & 36.95 & \textbf{65.88}
      & 26.35 & 63.65
      & 73.03 & 66.19
      & 48.95 & 82.81 & 62.98 \\
      \rowcolor{blue!5} Ours
      & \textbf{96.00} & \textbf{89.53}
      & 93.00 & \textbf{83.13}
      & 93.00 & \textbf{87.37}
      & \textbf{74.00} & 63.65
      & \textbf{37.04} & 64.43
      & \textbf{27.10} & \textbf{63.84}
      & \textbf{73.36} & \textbf{69.72}
      & \textbf{51.38} & \textbf{85.02} & 62.36 \\
      \midrule
      \multicolumn{18}{l}{\textbf{Stage III Post-Retrieval:} Resolving the Conflicts by Interference Elimination} \\
      w/o Elim
      & 94.00 & \textbf{89.87}
      & 89.50 & \textbf{85.63}
      & 88.00 & \textbf{87.78}
      & 68.00 & \textbf{67.54}
      & 35.29 & 63.00
      & \textbf{27.70} & 63.00
      & 70.84 & 64.68
      & 47.18 & 78.52 & 55.40 \\
      \rowcolor{blue!5} Ours
      & \textbf{96.00} & 89.53
      & \textbf{93.00} & 83.13
      & \textbf{93.00} & 87.37
      & \textbf{74.00} & 63.65
      & \textbf{37.04} & \textbf{64.43}
      & 27.10 & \textbf{63.84}
      & \textbf{73.36} & \textbf{69.72}
      & \textbf{51.38} & \textbf{85.02} & \textbf{62.36} \\
      \bottomrule
      \multicolumn{18}{l}{\scriptsize Note: Evaluation metric definitions are detailed in Sec.~\ref{sec:experimental_setup}. The symbol $\uparrow$ indicates that higher scores represent better performance.}
    \end{tabular}
  }
  }
\end{table*}

\section{Analysis and Discussion}

\subsection{Ablation Study}
\label{sec:ablation}
We conduct an ablation study on the Medical dataset to isolate the contributions of individual components in CodaRAG. The Medical dataset is selected for its relatively structured content and lower linguistic variability, enabling clearer attribution of observed performance differences. Table~\ref{tab:ablation_all} summarizes the overall results.
Our ablation follows the three stages of the pipeline. 
Stage I evaluates entry-only retrieval to isolate the effect of knowledge consolidation on initial entry quality without subsequent navigation. 
Stage II selectively deactivates individual association modules to assess their respective contributions. 
Stage III disables Interference Elimination to examine its impact on contextual reliability. 
To ensure a fair comparison, all ablation variants adopt a KG-only setting with limited entries, while keeping the contribution of Associative Navigation balanced.

\textbf{Stage I Indexing.}
Under entry-only settings, the consolidated KG outperforms the Base configuration on both retrieval and generation.
For example, retrieval recall (REC) on Creation queries improves from 35.84\% to 40.38\%, accompanied by a corresponding gain in generation accuracy (ACC).
This improvement suggests that the knowledge consolidation process produces higher-quality entries, reinforcing the importance of graph quality at the indexing stage~\cite{xiang2025use}, and providing a stronger starting point for subsequent navigation.
However, the remaining gap to the full system confirms that entry-only retrieval is insufficient without Associative Navigation to assemble dispersed evidence.

\textbf{Stage II Retrieval.}
Ablating individual association modules reveals complementary roles in evidence assembly.
Removing Semantic Association leads to the largest drop in retrieval recall (REC) and the greatest degradation in generation quality, underscoring the role of local expansion in surfacing relevant evidence~\cite{macavaney2020expansion}.
In contrast, removing Contextualized Association leads to moderate retrieval recall degradation, suggesting its role in identifying globally salient entities aligned with the query to organize dispersed evidence~\cite{tang2024multihop}. Although its removal slightly improves certain generation metrics (e.g., ACC), as observed in the Creation task, incorporating Contextualized Association improves coverage (COV) and faithfulness (FTH), leading to better overall generation quality.
Finally, removing Functional Association has limited impact on retrieval recall (REC) but degrades generation quality, with reductions in coverage (COV) and faithfulness (FTH) in summary and creation tasks, suggesting its role in retrieving structurally analogous entities that capture higher-level relational patterns.
Overall, these results suggest that the three pathways provide complementary retrieval evidence from different perspectives, supporting more effective evidence aggregation. While not uniformly improving all metrics, their combination yields more comprehensive evidence, as each pathway captures distinct aspects that are not fully represented by any single module, contributing differently to retrieval and generation.

\textbf{Stage III Post-Retrieval.}
Disabling Interference Elimination preserves retrieval recall but leads to clear drops in generation accuracy (ACC) and faithfulness (FTH).
This pattern suggests that Interference Elimination primarily regulates ambiguity rather than optimizing coverage, filtering noisy or competing evidence to support more coherent generation.
This observation indicates that irrelevant or distracting context can degrade generation quality~\cite{yoran2023making}.

In summary, the ablation study clarifies each stage’s distinct role: Stage I identifies relevant evidence units, Stage II connects dispersed evidence into an integrated support set, and Stage III refines the output for generation. Together, these results demonstrate the necessity of a staged pipeline across diverse query types.
We next turn to qualitative case studies to illustrate these mechanisms in practice.

\begin{table*}[t]
\caption{
Case studies illustrating generation outcomes. Key excerpts are presented here, with details provided in Appendix~\ref{sec:supplementary_case_study}. Colored text indicates entities identified by different association pathways in our method: \textcolor{freshblue}{Semantic Association}, \textcolor{freshred}{Contextualized Association}, and \textcolor{mygreen}{Functional Association}. The same color annotations are applied to baseline outputs only for alignment and comparison, and do not imply that these methods employ the corresponding mechanisms.
}
\vspace{-1em}
\label{tab:case_studies_compact}
\centering
\small
\renewcommand{\arraystretch}{1.1}
\setlength{\tabcolsep}{4pt}
\begin{tabular}{l p{0.86\textwidth}}
\toprule
\multicolumn{2}{p{0.98\textwidth}}{{\footnotesize \textbf{Medical Case:}} Discharge summary for \textit{PCNSL} transitioning from \textit{Induction} to \textit{Consolidation}.} \\
\cmidrule(lr){1-2}

{\footnotesize Ours} &
Completed induction with \textcolor{freshblue}{rituximab};
plans \textcolor{freshblue}{radiation}; 
monitored by \textcolor{freshred}{care team} using \textcolor{freshred}{imaging}
to manage \textcolor{mygreen}{nausea}. \\

{\footnotesize LightRAG} &
Completed induction with \textcolor{freshblue}{rituximab}; plans \textcolor{freshblue}{radiation};
transitioning to consolidation managed by \textcolor{freshred}{care team}
(missed specific symptom correlations). \\

{\footnotesize HippoRAG~2} &
Completed induction for PCNSL;
plan includes CBC monitoring and therapies like whole-brain \textcolor{freshblue}{radiation}
(lacks team coordination details). \\

\midrule

\multicolumn{2}{p{0.98\textwidth}}{{\footnotesize \textbf{Novel Case:}} Historical reasoning on Spanish descriptions in the progression affecting the Mayas.} \\
\cmidrule(lr){1-2}

{\footnotesize Ours} &
Spaniards (1517) described \textcolor{freshblue}{monuments};
despite iconoclasm, the Maya language survived,
preserving knowledge of figures like \textcolor{freshred}{Chaacmol}
(noted by scholar \textcolor{mygreen}{Brasseur de Bourbourg}). \\

{\footnotesize LightRAG} &
Spaniards (invaders) observed \textcolor{freshblue}{monuments} and fought Mayas;
despite destroying idols, the Maya language survived
(missed specific scholarly links). \\

{\footnotesize HippoRAG~2} &
Spaniards tried to wipe out ancient customs and Maya language
but failed (lacks details on monuments or figures). \\
\bottomrule
\end{tabular}
\end{table*}

\subsection{Case Study}

To illustrate how Associative Navigation improves generation fidelity, Table~\ref{tab:case_studies_compact} presents representative cases, with detailed retrieval traces and supporting evidence provided in Appendix~\ref{sec:supplementary_case_study}.

\textbf{Medical Case: Bridging Clinical Processes and Implicit Dependencies. } Medical guidelines often involve complex, non-adjacent logistical steps. For the PCNSL induction-to-consolidation query, LightRAG~\cite{guo2024lightrag} and HippoRAG~2~\cite{gutierrez2025rag} retrieve explicit therapies like \textit{rituximab} and \textit{radiation} but overlook the broader operational context or specific symptom correlations necessary for a clinical narrative. In contrast, CodaRAG leverages Contextualized Association to recover topologically distant but essential framework entities (e.g., \textit{care team}, \textit{imaging}). Furthermore, Functional Association bridges treatments to side effects like \textit{nausea}, providing a more holistic view of the patient’s clinical status that encompasses both disease control and quality-of-life impacts.
By linking these disparate elements, CodaRAG successfully navigates implicit clinical dependencies that extend beyond surface-level therapeutic entities.

\textbf{Novel Case: Resolving Granularity in Historical Narrative.}
Historical reasoning often struggles with the gap between broad temporal events and granular, verifiable evidence \cite{lin2025preliminary}. 
For the Spanish--Maya interaction query, LightRAG~\cite{guo2024lightrag} and HippoRAG~2~\cite{gutierrez2025rag} retrieve broad themes (e.g., \textit{monuments}) but fail to ground them in specific historical instances. 
In contrast, our method resolves this multiscale granularity bottleneck: Contextualized Association retrieves cultural figures such as \textit{Chaacmol}, while Functional Association validates claims via scholarly attributions to \textit{Brasseur de Bourbourg}. 
This establishes a structural division of labor where semantic retrieval provides the integrated narrative frame and advanced mechanisms supply authoritative, evidentiary granular evidence for historical precision. 

Overall, these cases show that effective RAG requires more than semantic relevance: it demands mechanisms that align retrieved evidence with the underlying reasoning intent through structured complementary associations.

\begin{table*}[t]
  \caption{Retrieval and generation performance on the Novel dataset with Mistral-medium-3.1~\cite{mistral_medium_3_1_2025}. 
All stages use the same model. 
Faithfulness for Creative Generation is marked as `-' due to invalid judge outputs.}
    \vspace{-1em}
  \label{tab:cross_model}
  \centering
  {\fontsize{7.5}{9}\selectfont
  \setlength{\tabcolsep}{4.5pt} 
  \renewcommand{\arraystretch}{1.15}
  \resizebox{\linewidth}{!}{
    \begin{tabular}{lcccccccc|ccccccccc}
      \toprule
      & \multicolumn{8}{c}{\textbf{Retrieval Performance $\uparrow$}}
      & \multicolumn{9}{c}{\textbf{Generation Performance $\uparrow$}} \\
      \cmidrule(lr){2-9}\cmidrule(lr){10-18}
      Method
        & \multicolumn{2}{c}{Fact}
        & \multicolumn{2}{c}{Reason}
        & \multicolumn{2}{c}{Summary}
        & \multicolumn{2}{c}{Creation}
        & \multicolumn{2}{c}{Fact}
        & \multicolumn{2}{c}{Reason}
        & \multicolumn{2}{c}{Summary}
        & \multicolumn{3}{c}{Creation} \\
      \cmidrule(lr){2-3} \cmidrule(lr){4-5} \cmidrule(lr){6-7} \cmidrule(lr){8-9} \cmidrule(lr){10-11} \cmidrule(lr){12-13} \cmidrule(lr){14-15} \cmidrule(lr){16-18}
      & REL & REC
      & REL & REC
      & REL & REC
      & REL & REC
      & RGL & ACC
      & RGL & ACC
      & COV & ACC
      & COV & FTH & ACC \\
      \midrule
      LightRAG
      & 91.41 & 91.24
      & 83.04 & 85.20
      & 83.37 & 88.45
      & \textbf{83.33} & 57.40
      & 5.55 & 20.18
      & 4.46 & 20.37
      & 75.91 & 21.57
      & 56.85 & - & 21.02 \\
      HippoRAG~2
      & 83.61 & 86.26
      & 74.84 & 74.75
      & 82.65 & 83.73
      & 52.77 & 55.07
      & \textbf{32.50} & 19.28
      & \textbf{25.76} & 21.22
      & 60.98 & 22.20
      & 51.30 & - & 21.13 \\
      \rowcolor{blue!5} Ours
      & \textbf{93.04} & \textbf{95.02}
      & \textbf{84.47} & \textbf{88.99}
      & \textbf{85.71} & \textbf{93.68}
      & 80.56 & \textbf{66.00}
      & 28.75 & \textbf{21.20}
      & 14.72 & \textbf{22.20}
      & \textbf{77.55} & \textbf{22.32}
      & \textbf{65.48} & - & \textbf{22.10} \\
      \bottomrule
      \multicolumn{18}{l}{\scriptsize Note: Evaluation metric definitions are detailed in Sec.~\ref{sec:experimental_setup}. The symbol $\uparrow$ indicates that higher scores represent better performance.} \\
    \end{tabular}
  }
 }
\end{table*}

\subsection{Cross-Model Robustness}
To assess cross-model robustness under a different backbone model, we evaluate performance on the Novel dataset, which contains unstructured narrative documents with high linguistic variability and diverse semantic expressions. While our main experiments use GPT-5-mini~\cite{openai_gpt5mini_2025}, we replace it with Mistral-medium-3.1~\cite{mistral_medium_3_1_2025} across graph construction, retrieval, generation, and evaluation to ensure a consistent and fair comparison across all stages of the pipeline.

As shown in Table~\ref{tab:cross_model}, our method ranks first across all question types, achieving the highest retrieval recall (REC) and generation accuracy (ACC), and outperforming LightRAG~\cite{guo2024lightrag} and HippoRAG~2~\cite{gutierrez2025rag}. These results demonstrate robustness to model changes and strong generalization across different LLM backbones, indicating that the effectiveness of our approach is not tied to a specific model choice. Additional results on the Medical dataset are provided in Appendix~\ref{sec:cross_supp}.



\section{Conclusion}
We introduced CodaRAG, a CLS-inspired graph-based RAG framework that treats retrieval as an active associative process. By combining Knowledge Consolidation, Associative Navigation, and Interference Elimination, CodaRAG constructs coherent evidence subgraphs that significantly enhance grounded generation. Evaluated on GraphRAG-Bench, our method achieves absolute gains of 7--10\% in retrieval recall and 3--11\% in generation accuracy across primary domains and task settings. 
These gains confirm CodaRAG's ability to navigate dispersed knowledge without succumbing to hyper-associative noise, demonstrating that \emph{connecting the dots with associativity} is an effective mechanism for overcoming evidence fragmentation in complex RAG scenarios.

\begin{acks}
This work was supported by the National Science and Technology Council (NSTC), Taiwan, for funding this research project under Grant No. NSTC 113-2740-H-002-001-MY3, “TAIHUCAIS: TAIwan HUmanities Conversational AI Knowledge Discovery System”.  
This work was also supported by the Ministry of Education (MOE) of Taiwan under the project Taiwan Centers of Excellence in Artificial Intelligence, through the NTU Artificial Intelligence Center of Research Excellence. 
Furthermore, we thank the National Center for High-performance Computing (NCHC) of National Applied Research Laboratories (NARLabs) in Taiwan for providing the necessary computational and storage resources.
\end{acks}


\bibliographystyle{ACM-Reference-Format}
\bibliography{myreference}

\newpage
\appendix

\section{Deep Dive Analysis into CodaRAG}

\subsection{Stage I Indexing: Knowledge Consolidation and Graph Structural Quality}
\label{sec:graph_quality}

\textbf{Structural Effects of Entity Discovery Methods.}
To evaluate the effect of \emph{Knowledge Consolidation} in Stage~I, we analyze how entity discovery paradigms shape graph structure when applied to a single document.
We compare Entity Type Discovery with Closed Information Extraction (ClosedIE) and Open Information Extraction (OpenIE) using a set of graph structural metrics, including number of nodes and edges, the average clustering coefficient (Avg.\ CC), the largest connected component ratio (LCC Ratio), the fragmentation ratio (Frag.\ Ratio), and the isolated entity ratio (Iso.\ Ratio).
As shown in Table~\ref{tab:graph_structure}, Entity Type Discovery yields substantially higher knowledge coverage, reflected by larger numbers of entities and relations. This richer coverage is accompanied by stronger local cohesion, indicated by a higher average clustering coefficient and a lower isolated entity ratio.
Although the largest connected component ratio is slightly lower than that of ClosedIE, the overall fragmentation remains comparable. This suggests a modest trade-off in global compactness for improved expressiveness and connectivity, consistent with the observed gains in downstream retrieval robustness.

\begin{table*}[htbp]
  \caption{Graph structural statistics under different entity discovery methods on a single document.}
    \vspace{-1em}
  \label{tab:graph_structure}
  \centering
  \small
  {\fontsize{7.5}{9}\selectfont
      \setlength{\tabcolsep}{10pt}
      \begin{tabular}{lcccccc}
        \toprule
        Method
        & \#Nodes $\uparrow$
        & \#Edges $\uparrow$
        & Avg.\ CC $\uparrow$
        & LCC Ratio $\uparrow$
        & Frag.\ Ratio $\downarrow$
        & Iso.\ Ratio $\downarrow$ \\
        \midrule
        ClosedIE
        & 727
        & 748
        & 0.1178
        & \textbf{0.695}
        & \textbf{0.0702}
        & 0.099 \\
        OpenIE
        & 691
        & 685
        & 0.1206
        & 0.631
        & 0.0955
        & 0.097 \\
        \rowcolor{blue!5} Ours
        & \textbf{893}
        & \textbf{880}
        & \textbf{0.1375}
        & 0.664
        & 0.0907
        & \textbf{0.073} \\
        \bottomrule
        \multicolumn{7}{l}{\scriptsize  Note: Evaluation Metric definitions are detailed in Sec.~\ref{sec:experimental_setup}. $\uparrow$ higher is better; $\downarrow$ lower is better.}
      \end{tabular}
      }
  \vskip -0.1in
\end{table*}

\begin{figure}[htbp]
  \centering
  \includegraphics[width=0.85\linewidth]{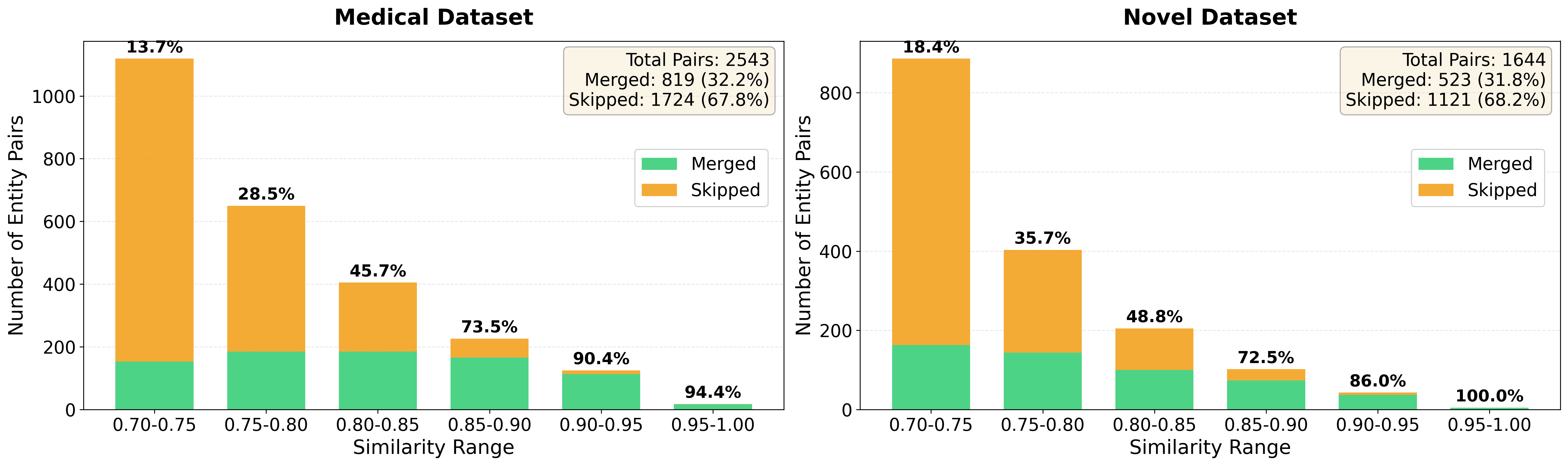}
  \vspace{-1em}
  \caption{
  Analysis of entity merging decisions under synonym proliferation.
  The figure illustrates representative cases where semantically equivalent or near-equivalent entity mentions are consolidated into canonical nodes, improving graph connectivity and reducing fragmentation while preserving evidence provenance.
  }
  \Description{
Distribution of entity pairs by similarity and merging decisions (merged vs skipped) across Medical and Novel datasets, showing higher merge rates at greater similarity levels.
}
  \label{fig:synonym_analysis}
\end{figure}

\textbf{Synonym Proliferation in LLM-based KG Construction.}
We analyze the synonym proliferation phenomenon in LLM-based entity extraction. As shown in Figure~\ref{fig:synonym_analysis}, LLM extraction yields many entity pairs with high embedding similarity, indicating extensive lexical variation and entity fragmentation. However, only a subset of these pairs are semantically suitable for merging, as high similarity does not necessarily imply true equivalence.
Across both the Medical and Novel datasets, approximately one-third of candidate pairs are judged merge-worthy, while the remainder correspond to distinct entities with meaningful semantic differences. This highlights a key challenge: although consolidation is necessary to reduce fragmentation, similarity alone is insufficient to prevent harmful over-merging. Naive similarity-based merging can therefore distort graph structure by collapsing semantically distinct entities. In contrast, Fragmented Entity Merging selectively consolidates compatible entities, improving connectivity and reducing fragmentation while preserving semantic granularity and evidence provenance.

\begin{table*}[t!]
\caption{Case study illustrating retrieval--generation division of labor across paradigms (Medical Case).
We report retrieval behaviors and the corresponding generation outcome. }
\label{tab:case_study_supp_1}
\centering
\small
\renewcommand{\arraystretch}{1.25}
\setlength{\tabcolsep}{6pt}
\resizebox{\linewidth}{!}{
\begin{tabular}{p{0.12\textwidth} p{0.90\textwidth}}
\toprule

\multicolumn{2}{p{0.96\textwidth}}{ 
\textbf{Medical Case:} Discharge summary for \textit{Primary Central Nervous System Lymphoma} transitioning from \textit{Induction} to \textit{Consolidation}.
} \\

\addlinespace[4pt]
\multicolumn{2}{c}{\bfseries Retrieval Trace} \\
\cmidrule(lr){1-2}

Ours &
\textbf{Entry:} Entities: Primary Central Nervous System Lymphoma; Induction; \dots\ ;
Relations: Primary Central Nervous System Lymphoma $\leftrightarrow$ Induction; Consolidation $\leftrightarrow$ Induction; \dots \newline
\textbf{Semantic Association:} Entities: Neurological Symptoms; Kidney Dysfunction; \dots\ ;
Relations: Primary Central Nervous System Lymphoma $\leftrightarrow$ Neurological Symptoms; High-Dose Methotrexate $\leftrightarrow$ Kidney Dysfunction; \dots \newline
\textbf{Contextualized Association:} Entities: Imaging; Care Team; \dots\ ;
Relations: Imaging $\leftrightarrow$ Biopsy; Imaging $\leftrightarrow$ Care Team; \dots \newline
\textbf{Functional Association:} Entities: Blood-Brain Barrier; Nausea; \dots
\\

\addlinespace[4pt]
LightRAG &
\textbf{Entry:} Entities: Consolidation Treatment; Induction Treatment; \dots\ ;
Relations: High-Dose Methotrexate $\leftrightarrow$ Induction Phase; Consolidation Phase $\leftrightarrow$ Primary CNS Lymphoma; \dots \newline
\textbf{One-hop:} Entities: Induction Phase; Radiation Therapy; \dots\ ;
Relations: Induction Treatment $\leftrightarrow$ Radiation Therapy; Induction Phase $\leftrightarrow$ Rituximab (Rituxan); \dots
\\

\addlinespace[4pt]
\multicolumn{2}{c}{\bfseries Generation (Detailed Summary)} \\
\cmidrule(lr){1-2}

Ours &
You completed induction with rituximab for primary CNS lymphoma and are transitioning to consolidation (e.g., radiation). Your multidisciplinary care team will coordinate this phase, using imaging and blood tests to monitor response. They will also manage side effects like nausea and fatigue to prioritize your quality of life.
\\

\addlinespace[2pt]
LightRAG &
You have completed induction chemotherapy for primary CNS lymphoma and are entering consolidation treatment to prevent relapse. Consolidation options may include radiation therapy or high-dose chemotherapy with stem cell rescue, depending on your condition and treatment response. Ongoing multidisciplinary care will focus on disease control and quality of life.
\\

\addlinespace[2pt]
HippoRAG~2 &
You have completed induction therapy for primary CNS lymphoma and are preparing for consolidation to reduce relapse risk. The plan includes therapies such as whole-brain radiation and ongoing monitoring via CBC and other lab tests. Immediate care is required for severe symptoms like fever.
\\

\bottomrule
\end{tabular}
}
\end{table*}

\begin{table*}[t!]
\caption{
Case study illustrating retrieval--generation division of labor across paradigms (Novel Case).
We report retrieval behaviors and the corresponding generation outcome.
}
\label{tab:case_study_supp_2}
\centering
\small
\renewcommand{\arraystretch}{0.9}
\setlength{\tabcolsep}{6pt}
\resizebox{\linewidth}{!}{
\begin{tabular}{p{0.12\textwidth} p{0.90\textwidth}}
\toprule

\multicolumn{2}{l}{
\textbf{Novel Case:} Historical reasoning on how the Spanish are described
in the progression affecting the Mayas.
} \\

\addlinespace[6pt]
\multicolumn{2}{c}{\bfseries Retrieval Trace} \\
\cmidrule(lr){1-2}

Ours &
\textbf{Entry:} Entities: Mayas; Maya Language; \dots\ ;
Relations: Mayas $\leftrightarrow$ Spanish Conquest; Maya Language $\leftrightarrow$ Spaniards; \dots \newline
\textbf{Semantic Association:} Entities: Monumental Inscriptions; Nahualt Invaders; \dots\ ;
Relations: Maya Language $\leftrightarrow$ Monumental Inscriptions; Maya People $\leftrightarrow$ Nahualt Invaders; \dots \newline
\textbf{Contextualized Association:} Entities: Macaw Totem (Moo); Chaacmol; \dots\ ;
Relations: Macaw Totem (Moo) $\leftrightarrow$ Chaacmol; Chaacmol $\leftrightarrow$ Chaacmol's Mausoleum; \dots \newline
\textbf{Functional Association:} Entities: Brasseur de Bourbourg; \dots
\\

\addlinespace[4pt]
LightRAG &
\textbf{Entry:} Entities: Spanish Conquest; Spaniards; \dots\ ;
Relations: Maya Language $\leftrightarrow$ Spaniards; Spaniards $\leftrightarrow$ Yucatan; \dots \newline
\textbf{One-hop:} Entities: Mayas; Maya Language; \dots\ ;
Relations: Maya Language $\leftrightarrow$ Mayas; Mayas $\leftrightarrow$ Yucatan; \dots
\\

\addlinespace[2pt]
\multicolumn{2}{c}{\bfseries Generation (Detailed Summary)} \\
\cmidrule(lr){1-2}

Ours &
The Spaniards (1517) were surprised to find well-built stone \textbf{monuments} and a civilized people. Although they invaded and destroyed many idols through iconoclasm, they could not destroy the Maya language; it preserved knowledge of figures like Chaacmol (noted by scholar Brasseur de Bourbourg).
\\

\addlinespace[2pt]
LightRAG &
The Spanish acted as intrusive colonial actors who observed Maya cities and monuments. They fought the Maya and attempted to erase indigenous customs by destroying idols. However, the Maya language survived, and the Conquest serves as a chronological marker for these events.
\\

\addlinespace[2pt]
HippoRAG~2 &
The Spanish attempted to completely wipe out the Mayas' ancient customs and language. They failed in this endeavor, as the language and many customs survived, forcing the Spanish to learn the indigenous tongue.
\\

\bottomrule
\end{tabular}
}
\end{table*}

\subsection{Stage II Retrieval: Associative Navigation and Retrieval Behavior}
\label{sec:supplementary_case_study}

\textbf{Retrieval-Generation Alignment.}
We present two case studies in Table~\ref{tab:case_study_supp_1} and Table~\ref{tab:case_study_supp_2} to illustrate how retrieved evidence shapes generation and influences the final output.
In the Medical Case, which requires a clinically structured discharge summary, our system constructs a comprehensive narrative by layering complementary evidence. Starting from core treatment anchors, the retrieval mechanism expands to include patient-specific safety factors via Semantic Association and care-process logistics through Contextualized Association. This is further reinforced by Functional Association, which integrates mechanistic context (e.g., blood--brain barrier constraints). Consequently, the generated summary accurately reflects the discharge genre by balancing transition rationale with monitoring protocols, whereas baselines like LightRAG~\cite{guo2024lightrag} tend to produce generic therapy descriptions or, in the case of HippoRAG~2~\cite{gutierrez2025rag}, procedurally complete text lacking specific evidence-based justification.
Similarly, in the Novel Case regarding the Spanish-Maya progression, our approach moves beyond static keyword matching to reconstruct a temporal narrative. By integrating cultural descriptors from Semantic Association with globally salient cues from Contextualized Association, the system successfully traces the evolution from initial contact and observation to conflict and linguistic persistence. Functional Association further stabilizes this account by aligning historical sources. This structural depth allows our model to preserve narrative specificity and temporal progression, in contrast to the compressed, high-level summaries produced by competing methods.
Overall, our qualitative advantage stems from the synergistic retrieval of attributes, process anchors, and background knowledge, yielding outputs that are not only factually grounded but also more precisely aligned with and strictly constrained to the query’s narrative or clinical requirements.

\begin{table*}[t]
\caption{Error analysis case studies summarizing queries, abstracted generated answers, and ground-truth evidences required for each query.
Evidences missing from the context are highlighted in \textbf{bold}.
}
\vspace{-1em}
\label{tab:error_case_study}
\centering
\small
\renewcommand{\arraystretch}{1.15}
\setlength{\tabcolsep}{6pt}

\begin{tabular}{p{0.60\textwidth} p{0.34\textwidth}}
\toprule

\multicolumn{2}{p{0.94\textwidth}}{
\textbf{Medical Case Query:}
Assume the role of a genetic counselor. Prepare a summary for a patient newly found to have a BRCA1 mutation, explaining cancer risk, surveillance, and preventive strategies.
} \\
\cmidrule(lr){1-2}

\textbf{Required Evidences} &
\textbf{Generation (Detailed Summary)} \\
\midrule

The patient has tested positive for a BRCA1 mutation; \newline
A BRCA1 mutation significantly increases breast cancer risk; \newline
Having a BRCA1 mutation is a strong risk factor; \newline
The mutation may impact family members; \newline
Enhanced surveillance is recommended; \newline
\textbf{Annual mammograms are recommended at an earlier age}; \newline
\textbf{Breast MRI is recommended at an earlier age}; \newline
\textbf{Preventive strategies include risk-reducing medication}; \newline
Preventive strategies include prophylactic surgery; \newline
\textbf{Bilateral mastectomy may be considered}; \newline
\textbf{Oophorectomy may be considered}; \newline
Discussion with oncology and surgical teams is recommended; \newline
Family members may consider genetic testing; \newline
Ongoing follow-up and support resources are available. &
States that a germline BRCA1 mutation increases breast and ovarian cancer risk, discusses genetic counseling, family testing, and general surveillance, and emphasizes coordination with a multidisciplinary care team without specifying concrete screening schedules or preventive interventions. \\

\midrule
\addlinespace[0.8em]

\multicolumn{2}{p{0.94\textwidth}}{
\textbf{Novel Case:}
How does the narrative link Mayab to the origins of the Mayas and their presence in Babylon?
} \\
\cmidrule(lr){1-2}

\textbf{Required Evidences} &
\textbf{Generation (Detailed Summary)} \\
\midrule

Mayab is a country of the Mayas; \newline
The Mayas are said to have originated from Mayab; \newline
\textbf{The Mayas established themselves in Babylon.} &
Identifies Mayab as the Maya homeland, compares Mayas with Chaldeans through cultural and architectural parallels, and introduces an Egypt-related migration hypothesis instead of explicitly stating Maya establishment in Babylon. \\

\bottomrule
\end{tabular}
\end{table*}

\textbf{Error Analysis: Impact of Missing Evidence.}
Table~\ref{tab:error_case_study} analyzes representative failure cases to examine how missing evidences in retrieval affect evidence recall and generation.
For each query, we compare the required ground-truth evidences with the generated output, highlighting evidences absent from the retrieved context.
In the Medical case, genetic risk information is retrieved, but guideline-level surveillance schedules and preventive interventions are missing, reducing evidence recall to 0.64 and resulting in a clinically plausible yet non-actionable summary.
In the Novel case, the absence of explicit evidence linking the Mayas to Babylon leads to partial recall (0.67), with generation substituting the missing claim with cultural analogies rather than unsupported assertions.
Across both cases, insufficient evidence granularity yields fluent but underspecified generation, indicating that these failures stem from retrieval limitations rather than hallucination.

\begin{table}[t]
\caption{Analysis of false deletions introduced by Interference Elimination.}
\vspace{-1em}
\label{tab:false_deletion}
\centering
\small
\setlength{\tabcolsep}{12pt}
\renewcommand{\arraystretch}{1.15}
\begin{tabular}{l c c}
\toprule
\textbf{Dataset}  & \textbf{Medical} &  \textbf{Novel} \\
\midrule
\textbf{False Elimination Rate (\%)} &   1.42 &  0.85  \\
\bottomrule
\end{tabular}
\end{table}

\subsection{Stage III Post Retrieval \& Answering: Interference Elimination and Error Propagation}
\label{sec:stage3_analysis}

To examine whether missing evidences could stem from overly aggressive filtering,
we conduct an attribution-based diagnostic analysis to quantify false deletions introduced by Interference Elimination.
Specifically, we identify gold evidences that are supported by the retrieved context before filtering,
and verify whether those same evidences remain supported afterward.
The sampled queries are evenly drawn across the four question types defined in the benchmark.
%
Table~\ref{tab:false_deletion} reports the false deletion rates across domains. Across 50 sampled queries per dataset, the false deletion rate remains consistently low, at 1.42\% on the Medical dataset and 0.85\% on the Novel dataset. These results indicate that Interference Elimination rarely removes evidence that was already supported prior to filtering, suggesting a generally conservative filtering behavior, although the risk of information loss cannot be completely eliminated.

\section{Supplementary Ablation Study on the Novel Dataset}
We further report an ablation study on the Novel dataset to examine whether the observed component-wise behaviors generalize to unstructured and linguistically diverse settings.
Table~\ref{tab:ablation_supp} presents retrieval and generation performance under the same systematically simplified configurations as in the Medical analysis.
The ablations follow the same stage-wise organization for systematic analysis.
Stage I evaluates the effect of knowledge consolidation by comparing entry-only retrieval against the Base configuration.
Stage II removes individual association modules to analyze their respective roles under high narrative variability.
Stage III disables Interference Elimination to assess its impact on evidence filtering and overall generation reliability.

\begin{table*}[!t]
  \caption{
    Ablation analysis of core components on the Novel dataset. 
    We conduct ablation studies by comparing our full framework with five variants: (1) Base (the original LightRAG~\cite{guo2024lightrag} baseline); (2) w/o SemA, w/o ConA, and w/o FunA, stripping the Semantic, Contextualized, and Functional Association modules; and (3) w/o Elim, excluding the Interference Elimination module. }
  \vspace{-1em}
  \label{tab:ablation_supp}
  \centering
  {\fontsize{7.8}{10}\selectfont
  \setlength{\tabcolsep}{4pt}
  \renewcommand{\arraystretch}{1.15}
  \resizebox{\linewidth}{!}{
    \begin{tabular}{lcccccccc|ccccccccc}
      \toprule
      & \multicolumn{8}{c}{\textbf{Retrieval Performance $\uparrow$}}
      & \multicolumn{9}{c}{\textbf{Generation Performance $\uparrow$}} \\
      \cmidrule(lr){2-9}\cmidrule(lr){10-18}
      Method
      & \multicolumn{2}{c}{Fact}
      & \multicolumn{2}{c}{Reason}
      & \multicolumn{2}{c}{Summary}
      & \multicolumn{2}{c}{Creation}
      & \multicolumn{2}{c}{Fact}
      & \multicolumn{2}{c}{Reason}
      & \multicolumn{2}{c}{Summary}
      & \multicolumn{3}{c}{Creation} \\
      \cmidrule(lr){2-3} \cmidrule(lr){4-5} \cmidrule(lr){6-7} \cmidrule(lr){8-9} \cmidrule(lr){10-11} \cmidrule(lr){12-13} \cmidrule(lr){14-15} \cmidrule(lr){16-18}
      & REL & REC
      & REL & REC
      & REL & REC
      & REL & REC
      & RGL & ACC
      & RGL & ACC
      & COV & ACC
      & COV & FTH & ACC \\
      \midrule
      \multicolumn{18}{l}{\textbf{Stage I Indexing:} Identifying the Dots by Knowledge Consolidation} \\
      Base
      & 85.99 & 69.22
      & 84.82 & 66.00
      & 85.46 & 61.30
      & \textbf{81.94} & 32.36
      & 38.02 & 54.64 
      & \textbf{25.98} & 43.54 
      & 49.87 & 43.07 
      & 35.28 & 71.89 & \textbf{44.37} \\
      \rowcolor{blue!5} Ours
      & \textbf{89.56} & \textbf{75.01}
      & \textbf{88.19} & \textbf{72.77}
      & \textbf{85.97} & \textbf{64.93}
      & 80.83 & \textbf{35.50}
      & \textbf{38.61} & \textbf{56.34} 
      & 24.50 & \textbf{45.61} 
      & \textbf{51.53} & \textbf{46.92} 
      & \textbf{37.12} & \textbf{77.90} & 40.50 \\
      \midrule
      \multicolumn{18}{l}{\textbf{Stage II Retrieval:} Connecting the Dots by Associative Navigation} \\
      w/o SemA
      & 91.30 & 78.03
      & 87.73 & 74.63
      & 91.07 & 73.69
      & 86.11 & 44.22
      & 39.85 & 57.89 
      & 26.10 & 45.69 
      & 57.59 & 47.61 
      & 46.74 & 76.37 & 43.74 \\
      w/o ConA
      & 94.05 & 82.22
      & 87.58 & 77.61
      & 92.60 & 76.84
      & \textbf{93.06} & 42.89
      & \textbf{40.43} & 58.85
      & 24.95 & 46.23 
      & 57.36 & 47.18 
      & 45.52 & 71.52 & 43.03 \\
      w/o FunA
      & 94.41 & 83.27
      & 89.26 & 76.98
      & 92.09 & 74.90
      & 87.50 & \textbf{50.16}
      & 39.77 & 58.04 
      & 25.86 & 46.08 
      & 58.94 & 46.96 
      & 43.31 & 74.43 & 42.56 \\
      \rowcolor{blue!5} Ours
      & \textbf{94.78} & \textbf{84.01}
      & \textbf{89.57} & \textbf{78.58}
      & \textbf{93.62} & \textbf{78.44}
      & 88.89 & 48.19
      & 40.42 & \textbf{59.41} 
      & \textbf{26.90} & \textbf{47.32} 
      & \textbf{59.87} & \textbf{48.52} 
      & \textbf{49.34} & \textbf{77.69} & \textbf{45.59} \\
      \midrule
      \multicolumn{18}{l}{\textbf{Stage III Post-Retrieval:} Resolving the Conflicts by Interference Elimination} \\
      w/o Elim
      & 94.60 & \textbf{85.50}
      & 89.26 & \textbf{81.65}
      & 93.37 & \textbf{78.47}
      & \textbf{88.89} & \textbf{50.41}
      & 38.45 & 59.30 
      & 25.85 & 46.31 
      & 56.45 & 48.08 
      & 45.06 & 75.94 & 43.78 \\
      \rowcolor{blue!5} Ours
      & \textbf{94.78} & 84.01
      & \textbf{89.57} & 78.58
      & \textbf{93.62} & 78.44
      & \textbf{88.89} & 48.19
      & \textbf{40.42} & \textbf{59.41} 
      & \textbf{26.90} & \textbf{47.32} 
      & \textbf{59.87} & \textbf{48.52} 
      & \textbf{49.34} & \textbf{77.69} & \textbf{45.59} \\
      \bottomrule
      \multicolumn{18}{l}{\scriptsize Note: Evaluation Metric definitions are detailed in Sec.~\ref{sec:experimental_setup}. The symbol $\uparrow$ indicates that higher scores represent better performance.} \\
    \end{tabular}
  }
  }
\end{table*}

\section{Supplementary Cross-Model Robustness on the Medical Dataset}
\label{sec:cross_supp}
We additionally report cross-model results on the Medical dataset to complement the main Novel analysis and examine robustness under a structured domain. Following the same protocol, we replace GPT-5-mini~\cite{openai_gpt5mini_2025} with Mistral-medium-3.1~\cite{mistral_medium_3_1_2025} for graph construction, retrieval, generation, and evaluation, ensuring that the entire pipeline is consistently executed under a single alternative model. 
As shown in Table~\ref{tab:cross_supp}, our method remains competitive across all question types, consistently achieving strong retrieval recall and stable generation accuracy, indicating that the benefits of associative navigation and interference-aware retrieval persist under model changes even in domain-specific settings.
\begin{table*}[t!]
  \caption{Retrieval and generation performance on the Medical dataset with Mistral-medium-3.1~\cite{mistral_medium_3_1_2025}. 
All stages use the same model. 
Faithfulness for Creative Generation is marked as `-' due to invalid judge outputs.}
    \vspace{-1em}
  \label{tab:cross_supp}
  \centering
  {\fontsize{7.5}{9}\selectfont
  \setlength{\tabcolsep}{4.5pt} 
  \renewcommand{\arraystretch}{1.15}
  \resizebox{\linewidth}{!}{
    \begin{tabular}{lcccccccc|ccccccccc}
      \toprule
      & \multicolumn{8}{c}{\textbf{Retrieval Performance $\uparrow$}}
      & \multicolumn{9}{c}{\textbf{Generation Performance $\uparrow$}} \\
      \cmidrule(lr){2-9}\cmidrule(lr){10-18}
      Method
        & \multicolumn{2}{c}{Fact}
        & \multicolumn{2}{c}{Reason}
        & \multicolumn{2}{c}{Summary}
        & \multicolumn{2}{c}{Creation}
        & \multicolumn{2}{c}{Fact}
        & \multicolumn{2}{c}{Reason}
        & \multicolumn{2}{c}{Summary}
        & \multicolumn{3}{c}{Creation} \\
      \cmidrule(lr){2-3} \cmidrule(lr){4-5} \cmidrule(lr){6-7} \cmidrule(lr){8-9} \cmidrule(lr){10-11} \cmidrule(lr){12-13} \cmidrule(lr){14-15} \cmidrule(lr){16-18}
      & REL & REC
      & REL & REC
      & REL & REC
      & REL & REC
      & RGL & ACC
      & RGL & ACC
      & COV & ACC
      & COV & FTH & ACC \\
      \midrule
      LightRAG
      & \textbf{92.00} & 91.57
      & 86.00 & 89.49
      & \textbf{90.00} & 91.84
      & \textbf{77.50} & 84.41
      & 3.85 & 20.87
      & 5.12 & 21.08
      & 84.26 & 22.02
      & 59.60 & - & 22.61 \\
      HippoRAG~2
      & 84.00 & 90.64
      & 79.50 & 89.33
      & 89.50 & 89.37
      & 68.00 & 83.75
      & \textbf{33.52} & 20.69
      & \textbf{32.91} & 21.85
      & 80.76 & 22.90
      & 62.51 & - & 22.60 \\
      \rowcolor{blue!5} Ours
      & 91.00 & \textbf{92.13}
      & \textbf{87.00} & \textbf{90.10}
      & 87.50 & \textbf{92.65}
      & 72.50 & \textbf{85.39}
      & 25.13 & \textbf{21.59}
      & 17.66 & \textbf{23.77}
      & \textbf{85.33} & \textbf{23.16}
      & \textbf{63.69} & - & \textbf{22.73} \\
      \bottomrule
      \multicolumn{18}{l}{\scriptsize Note: Evaluation Metric definitions are detailed in Sec.~\ref{sec:experimental_setup}. The symbol $\uparrow$ indicates that higher scores represent better performance.} \\
    \end{tabular}
  }
  }
\end{table*}

\section{Computational Complexity and Cost Discussion}
\label{sec:cost_analysis}

CodaRAG introduces additional computational overhead via LLM components.
While not primarily optimizing for latency or throughput, our design explicitly prioritizes
retrieval quality, faithfulness, and robustness in complex task settings.
Below, we discuss the computational considerations of CodaRAG across indexing and inference phases, and clarify the intended design trade-offs.

\textbf{Indexing Costs: One-time Consolidation.}
\emph{Entity Type Discovery} and \emph{Fragmented Entity Merging} rely on LLM calls, leading to higher upfront indexing costs than naive extraction-based pipelines. These costs are incurred only once during graph construction and amortized over subsequent queries, making the per-query indexing overhead negligible in practice.

\textbf{Inference Costs: Controlled Retrieval.}
At inference time, \emph{Interference Elimination} introduces additional computation to regulate associative retrieval, reflecting a deliberate trade-off that prioritizes retrieval quality and generation reliability over raw inference speed. While this increases inference overhead, it consistently improves generation faithfulness and accuracy, particularly in high-stakes domains such as medical question answering, where unreliable outputs are costly.

Overall, CodaRAG does not primarily optimize for computational efficiency, but rather aims to investigate how structured, interference-aware retrieval influences reasoning reliability. Future work may explore more efficient variants for cost-sensitive and large-scale deployments.


\begin{table*}[ht]
\centering
\small
\caption{Summary of Evaluation Metrics for Retrieval, Generation, and Graph Structure.}
\label{tab:metrics_summary}
\vspace{-1em}
\renewcommand{\tabularxcolumn}[1]{m{#1}} 
\begin{tabularx}{\textwidth}{llcX}
\toprule
\textbf{Category} & \textbf{Metric} & \textbf{Formula} & \textbf{Description \& Key Intuition} \\
\midrule

\multirow{4}{*}{\textbf{Retrieval}} 
& Context Relevance 
& $\frac{r_1 + r_2}{2}$ 
& Measures how relevant the retrieved context is to the query based on annotator scores $r_i \in \{0, 0.5, 1\}$. \\
\cmidrule{2-4}
& Evidence Recall 
& $\frac{1}{|R|} \sum_{c \in R} \mathbb{I}(S(c, C))$ 
& Evaluates the proportion of reference claims supported by the retrieved context. \\
\midrule

\multirow{8}{*}{\textbf{Generation}} 
& Rouge-L
& $\frac{(1+\beta^2) P_{LCS} R_{LCS}}{R_{LCS} + \beta^2 P_{LCS}}$ 
& Captures lexical overlap between the generated and reference answers using the longest common subsequence (LCS). \\
\cmidrule{2-4}
& Answer Accuracy 
& $\alpha \mathrm{FC} + (1-\alpha)\mathrm{SS}$ 
& Combines factual correctness (FC) and semantic similarity (SS) with $\alpha=0.5$. \\
\cmidrule{2-4}
& Faithfulness 
& $\frac{|\{c \in A \mid S(c, C)=1\}|}{|A|}$ 
& Reflects the proportion of generated claims supported by the retrieved context, indicating hallucination avoidance. \\
\cmidrule{2-4}
& Evidence Coverage 
& $\frac{|\{k \in K \mid M(k, G)=1\}|}{|K|}$ 
& Assesses how completely the required evidence is reflected in the generated answer. \\
\midrule

\multirow{6}{*}{\textbf{Graph}} 
& Avg. Clustering Coeff. 
& $\frac{1}{|E|} \sum_{e \in E} C(e)$ 
& Indicates the level of local connectivity and triadic closure in the knowledge graph. \\
\cmidrule{2-4}
& LCC Ratio 
& $|\mathrm{LCC}| / |E|$ 
& Quantifies the dominance of the largest connected component in the graph. \\
\cmidrule{2-4}
& Isolated Entity Ratio 
& $|\mathrm{Iso}| / |E|$ 
& Measures the proportion of nodes with no connections. \\
\cmidrule{2-4}
& Fragmentation Ratio 
& $NC / |E|$ 
& Captures structural fragmentation beyond isolated nodes. \\
\bottomrule
\end{tabularx}
\end{table*}

\section{Evaluation Metrics}
\label{sec:eval_metrics}

We evaluate retrieval and generation using metrics from GraphRAG-Bench~\cite{xiang2025use}.
Specifically, let $R$ denote the set of reference claims, $C$ the retrieved context, and $G$ the generated answer. 
$S(c, C) \in \{0,1\}$ indicates whether a claim $c$ is supported by $C$, and $\mathbb{I}(\cdot)$ is the indicator function. 
For generation evaluation, $A$ denotes the set of atomic claims extracted from $G$, 
$K$ denotes the set of required evidence elements, and 
$M(k, G) \in \{0,1\}$ indicates whether a required evidence element $k$ is reflected in $G$. 
For Rouge-L, $P_{LCS}$ and $R_{LCS}$ denote precision and recall computed from the longest common subsequence (LCS), and $\beta$ controls their relative weights.
Answer Accuracy is defined as a weighted sum $\alpha \mathrm{FC} + (1-\alpha)\mathrm{SS}$, where we set $\alpha=0.5$. 
Here, $\mathrm{FC} = \frac{2TP}{2TP + FP + FN}$ evaluates factual correctness based on claim-level matching, 
where $TP$, $FP$, and $FN$ denote true positives, false positives, and false negatives at the claim level. 
$\mathrm{SS} = \cos(f_G, f_R)$ evaluates semantic similarity, where $f_G$ and $f_R$ are the semantic embeddings of the generated and reference answers.
For graph analysis, $\mathcal{G} = (E, \mathcal{R})$ denotes the knowledge graph, where $E$ is the set of entities. 
$C(e)$ denotes the clustering coefficient of entity $e$.
Let $\mathrm{LCC}$ denote the set of nodes in the largest connected component, 
$\mathrm{Iso}$ the set of isolated entities, and $NC$ the number of non-isolated connected components.

As summarized in Table~\ref{tab:metrics_summary}, our evaluation comprehensively covers three complementary aspects: \textbf{Retrieval Quality}, which assesses whether the retrieved context is relevant and contains sufficient supporting evidence; \textbf{Generation Quality}, which evaluates factual grounding, semantic correctness, lexical overlap, and completeness while accounting for hallucination; and \textbf{Graph Structure}, which analyzes the connectivity, local structure, and fragmentation of the constructed graph, along with their impact on multi-hop reasoning.

\section{Benchmark Selection}
\label{sec:benchmark}
We focus on GraphRAG-Bench~\cite{xiang2025use}, as it is explicitly designed to evaluate graph-based RAG systems and better reflects the effects of graph construction and structured retrieval.
In contrast, widely used benchmarks such as HotpotQA~\cite{yang2018hotpotqa} and 2WikiMultiHopQA~\cite{ho2020constructing} assess retrieval primarily through alignment with annotated supporting evidence at the passage or sentence level. In our setting, information is consolidated into entity and relation representations during graph construction rather than preserved as original text spans, making direct alignment with annotated evidence less straightforward. Consequently, evaluating structured evidence requires assessment beyond text-span alignment, as supported by the evaluation framework of GraphRAG-Bench~\cite{xiang2025use}. 
Due to this representation mismatch, semantically correct evidence retrieved in structured form may not always correspond to annotated text units, which can affect evaluation under text-span-based criteria. We acknowledge that excluding standard QA benchmarks may limit direct comparability, as discussed in the main text.

\section{Experimental Hyperparameter Settings}
\label{sec:appendix_hyperparameters}

This appendix summarizes the hyperparameter configurations used in all experiments, including the main experiments and ablation studies.
All methods are evaluated under the same data splits and evaluation protocols, with method-specific retrieval parameters set according to either prior work or controlled experimental design.
For all experiments, documents are segmented using a fixed chunk size of 1200 tokens with an overlap of 100 tokens to ensure consistent context granularity across methods. Unless otherwise specified, experiments using GPT-5-mini~\cite{openai_gpt5mini_2025} are conducted with the \texttt{reasoning\_effort} parameter set to low.

\textbf{FastRP Configuration.}
FastRP~\cite{chen2019fast} is used to encode structural similarity among entities, providing an efficient and unsupervised structural signal for retrieval-time Functional Association. Embeddings are computed with a dimension of 256, and a negative normalization strength ($-0.1$) is applied to mitigate high-degree dominance and stabilize representations. Iteration weights $\{1.0, 1.0, 0.5, 0.25\}$ emphasize closer neighborhoods while retaining sensitivity to more distant structural patterns. Concretely, entity embeddings $\mathbf{X}$ are computed as
\begin{equation}
\label{eq:fastrp_combined}
\mathbf{X} = \sum_{k=0}^{K} w_k \, \mathbf{D}^{\,r} \mathbf{S}^{\,k} \mathbf{R}, \quad \text{where} \quad \mathbf{S} = \mathbf{D}^{-\frac{1}{2}} \left(\mathbf{A}_{\text{weight}} + \mathbf{I}\right) \mathbf{D}^{-\frac{1}{2}},
\end{equation}
where $K$ denotes the maximum propagation depth (set to $K=3$), $\mathbf{R} \in \mathbb{R}^{|\mathcal{E}| \times d}$ is an initial random projection matrix whose row vectors are L2-normalized, $w_k$ controls the contribution of the $k$-hop neighborhood, and $r$ is a degree normalization parameter (set to $r=-0.1$). The matrix $\mathbf{A}_{\text{weight}}$ denotes a co-occurrence–weighted adjacency matrix over entity relations, $\mathbf{I}$ is the identity matrix, and $\mathbf{D}$ is the corresponding degree matrix. Edge weights reflect corpus-level co-occurrence statistics to emphasize structurally informative relations.

\begin{table}[t!]
\caption{Hyperparameters used in the main experiments.}
\vspace{-1em}
\centering
\small
\resizebox{\linewidth}{!}{
\setlength{\tabcolsep}{4pt}
\renewcommand{\arraystretch}{0.9}
\begin{tabular}{l l c p{9cm}}
\toprule
\textbf{Method} & \textbf{Hyperparameter} & \textbf{Value} & \textbf{Description} \\
\midrule
LightRAG 
& \texttt{top\_k} & 10 
& - \\

& \texttt{chunk\_top\_k} & 5 
& - \\
\midrule
HippoRAG~2
& \texttt{retrieval\_top\_k} & 10 
& - \\

& \texttt{linking\_top\_k} & 20 
& - \\

& \texttt{qa\_top\_k} & 5 
& - \\
\midrule
Ours
& \texttt{top\_k} & 10 
& Number of initial entity and relation entries used to initiate associative retrieval \\

& \texttt{chunk\_top\_k} & 5 
& Number of text chunks provided to the generator \\

& \texttt{top\_neighbors} & 10 
& Maximum neighbors explored per node during \emph{Semantic Association} \\

& \texttt{top\_ppr\_nodes} & 20 
& Nodes selected by PPR in \emph{Contextualized Association} \\

& \texttt{top\_fastrp\_nodes} & 10 
& Structurally analogous nodes selected in \emph{Functional Association} \\
\bottomrule
\end{tabular}
\label{tab:main_hyperparameters}
}
\end{table}

\begin{table}[t!]
\caption{Hyperparameters used in the ablation studies.}
\vspace{-1em}
\centering
\resizebox{\linewidth}{!}{
\small
\setlength{\tabcolsep}{4pt}
\renewcommand{\arraystretch}{0.9}
\begin{tabular}{l l c p{9cm}}
\toprule
\textbf{Method} & \textbf{Hyperparameter} & \textbf{Value} & \textbf{Description} \\
\midrule
Ours (Ablation)
& \texttt{top\_k} & 3 
& Number of initial entity and relation entries used to initiate associative retrieval \\

& \texttt{chunk\_top\_k} & 0 
& Number of text chunks provided to the generator \\

& \texttt{top\_neighbors} & 8 
& Maximum neighbors explored per node during \emph{Semantic Association} \\

& \texttt{top\_ppr\_nodes} & 15 
& Nodes selected by PPR in \emph{Contextualized Association} \\

& \texttt{top\_fastrp\_nodes} & 15 
& Structurally analogous nodes selected in \emph{Functional Association} \\
\bottomrule
\end{tabular}
\label{tab:ablation_hyperparameters}
}
\end{table}

\textbf{Main Experiment Settings.}
Instead of enforcing a fixed token budget, we standardize the retrieval scope by aligning the number of retrieved chunks or graph items with the native operating conditions of each method.
While LightRAG~\cite{guo2024lightrag} and our method retrieve both KG evidence and text chunks, in contrast to the text-only retrieval of HippoRAG~2~\cite{gutierrez2025rag}, these differences reflect their respective retrieval paradigms rather than arbitrary design choices.
Hyperparameters in our method are selected via a coarse grid search on a held-out subset of the data based on retrieval and generation metrics, and are fixed across all datasets and tasks (see Table~\ref{tab:main_hyperparameters}).
As a result, performance differences are primarily attributable to the quality and structural coherence of the retrieved evidence, rather than variations in context length.

\textbf{Ablation Study Settings.}
Ablation studies are conducted under a more constrained retrieval budget to isolate the effects of individual components.
In particular, text chunk augmentation is disabled and the number of entry dots is intentionally limited, enabling evaluation of graph-only retrieval behavior under controlled conditions.
Retrieval budgets are configured to keep the contributions of different associative modules balanced, so that observed differences primarily reflect the characteristics of each association mechanism rather than disparities in retrieval effort.
Table~\ref{tab:ablation_hyperparameters} summarizes the hyperparameters used in the ablation experiments.

\section{Prompt Specifications}
We employ prompts at different stages of the pipeline: in Stage I, prompts support Entity Type Discovery, Information Extraction, and Fragmented Entity Merging to construct a consolidated knowledge structure; in Stage II, they guide query-related cue generation as the first step of Entry Finding in Associative Navigation; and in Stage III, they support Interference Elimination followed by response generation based on the refined context.

\subsection{Stage I Indexing}

\subsubsection{Entity Type Discovery}
\label{sec:discovery_prompt}

We adopt an Entity Type Discovery module to induce domain-appropriate entity types directly from document content. This process consists of two steps: type suggestion and type refinement. Candidate types are first proposed based on document content and structural patterns, and then refined to reduce redundancy while preserving domain-specific distinctions.

\begin{codelist}
--- Type Suggestion ---
1. Analyze document content, structural elements, and domain-specific patterns.
2. Identify recurring entity categories, their contextual roles, and relationships.
3. Propose non-overlapping and domain-appropriate entity types that improve extraction coverage with concise explanations.

--- Type Refinement ---
1. Identify duplicate or highly overlapping entity types based on semantic similarity.
2. Consolidate redundant types while preserving meaningful domain-specific distinctions.
3. Produce a concise and well-balanced schema that reduces redundancy while maintaining coverage.
\end{codelist}

\subsubsection{Information Extraction}
Following LightRAG~\cite{guo2024lightrag}, we adopt a structured prompt to extract entities and binary relationships from text.
\begin{codelist}
1. Identify entities from the text based on predefined entity types and assign consistent names and descriptions.
2. Extract direct relationships among identified entities and decompose complex interactions into binary relations.
3. Produce structured outputs with entity and relation descriptions grounded strictly in the input text.
\end{codelist}

\subsubsection{Fragmented Entity Merging}
\label{sec:merging_prompt}

We employ a pairwise evaluation to determine whether two entities refer to the same real-world entity. Candidate pairs are first filtered by embedding similarity, followed by LLM-based decisions.

\begin{codelist}
1. Compare entity identifiers and descriptions to assess semantic equivalence.
2. Enforce strict matching criteria by prioritizing proper noun consistency and avoiding merges based on superficial similarity.
3. Perform merging only when there is high confidence they refer to the same real-world entity; otherwise keep them distinct.
\end{codelist}

\subsection{Stage II Retrieval and Stage III Post-Retrieval}

\subsubsection{Query-related Cues Generation}
Following LightRAG~\cite{guo2024lightrag}, we extract high-level and low-level keywords from the user query. In our framework, these keywords serve as query-related cues that condition associative retrieval.

\begin{codelist}
1. Extract high-level keywords that capture the overall query intent and semantic scope.
2. Identify low-level keywords corresponding to specific entities, terms, or detailed aspects.
3. Produce concise and meaningful keyword sets derived strictly from the query for retrieval guidance.
\end{codelist}

\subsubsection{Response}
We use grounded prompts to generate answers from retrieved entities, relations, and document chunks while maintaining strict support from the provided context.

\begin{codelist}
1. Identify relevant entities and relations from the provided context to determine the core semantic structure.
2. Ground the selected information using supporting document chunks from the context, preserving original phrasing when possible.
3. Generate a precise answer that integrates only necessary facts, ensuring all statements are strictly supported by the context.
\end{codelist}

\subsubsection{Interference Elimination}
\label{sec:IE_prompt}
We leverage LLMs to refine retrieved evidence by suppressing irrelevant or ambiguous items while preserving potentially useful supporting information.
\begin{codelist}
1. Evaluate retrieved entities and relations to identify items that do not contribute to the answer.
2. Suppress irrelevant or ambiguous evidence while preserving potentially useful intermediate or supporting information.
3. Retain a coherent and query-aligned evidence set by removing distracting elements without breaking necessary relational connections.
\end{codelist}

\end{document}